\documentclass[lettersize,journal]{IEEEtran}
\usepackage{amsmath,amsfonts,amsthm}
\usepackage{amssymb}
\usepackage[linesnumbered, ruled, vlined]{algorithm2e} 
\usepackage[english]{babel}
\SetKwInOut{Input}{Input}
\SetKwInOut{Output}{Output}
\SetKw{MyAnd}{and}
\SetKw{Break}{break}
\SetKw{Continue}{continue}
\usepackage{array}
\usepackage[caption=false,font=normalsize,labelfont=sf,textfont=sf]{subfig}
\usepackage{textcomp}
\usepackage{stfloats}
\usepackage{url}
\usepackage{verbatim}
\usepackage{graphicx}
\usepackage{cite}
\usepackage{xcolor}
\usepackage{multirow}
\usepackage[normalem]{ulem}
\usepackage{array}

\begin{document}
\title{Min-Max Regret Task Allocation and Planning of Heterogeneous Multi-Robot System in Partially Known Environments}

\author{Xinkai Liang, Huixuan Chan, Ying Liu, Yangxi Shi and Hao Fang, \textit{Member, IEEE}
\thanks{The authors are with the School of Automation, Beijing Institute of Technology, Beijing 100081, China. (e-mail: Xinkai Liang, yudubai@bit.edu.cn; Hao Fang, fangh@bit.edu.cn)}%
\thanks{This work was supported in part by the National Natural Science Foundation of China under Grant No. 62133002, and in part by the Beijing Natural Science Foundation under Grant No. L259052.}
}

\markboth{Journal of \LaTeX\ Class Files,~Vol.~14, No.~8, August~2021}%
{Shell \MakeLowercase{\textit{et al.}}: A Sample Article Using IEEEtran.cls for IEEE Journals}


\maketitle

\begin{abstract}
Efficient task allocation for large-scale Heterogeneous Multi-Robot Systems (HMRS) is critical, yet dealing with complex temporal logic tasks in partially known environment (PKE) remains a computational bottleneck. Existing approaches often struggle to balance exploring uncertain regions and exploiting known resources, while also suffering from exponential computational complexity. To address these issues, this paper presents a robust planning framework that simultaneously handles high-level logical constraints and environmental uncertainty without sacrificing scalability. We formulate the problem as a min-max regret optimization, proposing a Region-Binding Atomic Proposition (RbAP) to capture resource uncertainty within the automaton structure. To solve this, we propose the Extended Planning Decision Tree (E-PDT) equipped with a novel Regret-based Branch-and-Bound (BnB) strategy. Unlike traditional methods that rely on prior probabilities or worst-case analysis, our approach dynamically prunes suboptimal policies, effectively balancing the need for information gathering (exploration) and task completion (exploitation). Theoretical analysis confirms the feasibility and completeness of our approach. Extensive numerical and physical experiments demonstrate that the proposed framework achieves near-linear scalability with respect to the number of robots and types, significantly outperforming MILP-based baselines in both solution quality and computational efficiency.
\end{abstract}

\renewcommand{\abstractname}{Note to Practitioners}
\begin{abstract}
Task allocation for large-scale heterogeneous multi-robot systems is critical in applications like warehouse logistics and disaster response. However, existing planning methods often struggle with environmental uncertainty, relying on unavailable probability data or failing to scale. This work presents a fast, robust planning framework that uses a regret-minimization strategy to balance exploring uncertain areas with completing tasks. Unlike optimization-based approaches that can be computationally prohibitive, our E-PDT method achieves near-linear scalability, coordinating thousands of robots in seconds. This allows practitioners to deploy large robot fleets that adaptively search for resources without needing precise prior knowledge. Current limitations assume a known static map topology with uncertain resource locations; future extensions will address fully dynamic and unknown environments.
\end{abstract}


\section{Introduction}
\IEEEPARstart{R}{ecent}
advancements in autonomous robots have enabled multi-robot system to perform task allocation and path planning within unstructured environments \cite{lavalle2006planning, garrett2021integrated}. In this context, Linear Temporal Logic (LTL) has gathered significant attention as a formal language capable of expressing rich task specifications \cite{luckcuck2019formal, belta2019formal}. It can effectively articulate constraints for long-range navigation \cite{belta2007symbolic} and human-robot collaboration \cite{wells2021finite}, as well as more detailed tasks such as persistent surveillance of specific regions \cite{chen2024optimal} and visiting distinct locations in a prescribed logical sequence \cite{ulusoy2014receding}. In such applications, HMRS, composed of robots with diverse sensing and manipulation capabilities, have become increasingly preferred \cite{yu2021distributed}. To address the computational complexity inherent in these complex systems, extensive research has been conducted by numerous scholars \cite{guo2016task, schillinger2018simultaneous, luo2022temporal, leahy2022fast, chen2025real}. However, when traversing real-world environments that are not fully known, it is essential to take environmental information into account during the planning phase to generate feedback strategies capable of dynamic response. Although Markov Decision Processes (MDPs) are typically employed in such scenarios to quantify uncertainty \cite{lacerda2019probabilistic, guo2018probabilistic, cai2021optimal}, this approach relies on prior probabilities and is often infeasible when exploring an environment for the first time. Conversely, relying solely on the worst-case analysis tends to yield overly conservative strategies. Consequently, \cite{blackwell1956analog} et al. proposed regret as an alternative metric. Although this metric has seen some applications in autonomous robots \cite{zhao2025no}, existing methods suffer from high computational complexity, preventing their scalability to large-scale HMRS.

\subsection{Related Works}

To address the challenges of environmental uncertainty and computational complexity, existing research has evolved along two complementary layers: planning and decision-making algorithms, as well as task representation and formalization.

Planning and decision-making algorithms in partially known or unknown environments can be categorized into reactive and non-reactive approaches. Reactive planning, which can be environment-centric, task-centric, robot-centric, or a combination thereof, while environment-centric is the focus of this section. For instance, \cite{ayala2013temporal} employed interleaved automaton search and exploration, using a real-time monitor to evaluate the optimal exploration direction. The work of \cite{guo2015multi} proposed distributed, real-time plan reconfiguration, where robots exchange knowledge via sensing and local communication, utilizing an on-the-fly product automaton for efficient plan revision. In \cite{lahijanian2016iterative}, an iterative replanning framework was introduced; when new obstacles appear, it introduces three partial satisfaction methods to ensure safety while maximizing the satisfaction of soft constraints. Dynamic disturbances in the environment were modeled as an adversarial game problem to synthesize reactive strategies in \cite{he2017reactive}. The work of \cite{kantaros2020reactive} utilized online decomposition of global LTL tasks into single-robot reachability problems, performing reactive replanning based on maps continuously updated by SLAM. While non-reactive planning approaches often model the environment probabilistically. For instance, model-free reinforcement learning has been employed to maximize the satisfaction probability of LTL specifications in \cite{hasanbeig2019reinforcement, bozkurt2020control, cai2020learning}. The work of \cite{cai2023overcoming} further combined RRT* with deep reinforcement learning (DRL) to address exploration in cluttered environments. Diverging from maximizing satisfaction probability, \cite{zhao2023explore, zhao2025no} innovatively adopted regret as the planning metric to address PKE. 

Despite the progress in planning and decision-making algorithms for handling uncertainty, a critical bottleneck remains: the exponential growth of the state space as the number of robots, tasks, or environmental regions increases. Since formal methods were first applied to verify task planning \cite{baier2008principles}, a significant body of research has focused on developing algorithms to reduce computational complexity. The work of \cite{saha2014automated} simplified the problem by predefining motion primitives and performing a hierarchical search over their compositions. In \cite{tumova2016multi}, the problem was decomposed into a series of finite-horizon problems for iterative solving. \cite{schillinger2018decomposition} significantly reduced complexity by decomposing LTL specifications into a set of independently executable task specifications. Random search and cross-entropy optimization were used to determine the optimal robots for task completion in \cite{banks2020multi}. A series of studies in \cite{sampling-based, kantaros2020stylus, abstraction-free} integrated temporal logic constraints into a sampling-based optimal control framework, incrementally building optimal paths by intelligently guiding the sampling direction. The work of \cite{bai2022hierarchical} proposed a hierarchical framework to avoid constructing the full product automaton. More recently, \cite{liu2024fast} used relaxed partially ordered sets (po-sets) instead of Buchi automata and employed an incremental algorithm to avoid global synchronous products. Despite this progress, these methods still struggle to scale of problems involving thousands of robots. The work of \cite{chen2024pdt} departed from traditional product automata, instead proposing Planning Decision Tree (PDT) that represents task progression, demonstrating scalability to 10,000 robots across 100 different categories, which is extended to reactive planning \cite{chen2025real}. 
In their following works, Li et al. \cite{li2026formal} proposed a reactive multiconstraint planning decision tree (RMC-PDT) to address heterogeneous task allocation in uncertain semantic environments. They \cite{li2026environment} further introduced an LLM-guided framework that utilizes dual-system temporal logics to autonomously infer and allocate tasks based on environmental resource changes. However, while they account for environmental semantic uncertainty, their work primarily emphasizes reactive replanning rather than proactively avoiding risks during the initial planning stage.

Beyond reducing computational complexity, many researchers have also explored the semantic expressiveness of temporal logic languages. For instance, Signal Temporal Logic (STL) is designed to express continuous-time signals \cite{maler2004monitoring}. For discrete-time signals, representative examples include Truncated Linear Temporal Logic (TLTL) \cite{li2018policy}, Metric Interval Temporal Logic (MITL) \cite{alur1996benefits}, and Time-Window Temporal Logic (TWTL) \cite{vasile2017time}. Beyond expressing temporal constraints, cLTL \cite{sahin2019multirobot} effectively captures quantity-based multi-robot tasks, while scLTL \cite{kloetzer2016multi} enables planners to generate finite-length plans for task constraints that do not require infinite trajectories. In terms of extending predicate definitions, Li et al. \cite{li2023fast} captures the compatibility, mutual exclusion, and irrelevance relationships between tasks, and considering time variable capabilities of robots in \cite{li2025task}. Chen et al. \cite{chen2024fast} expresses spatial, temporal, and sequential constraints on task execution. However, none of the aforementioned methods address the semantic expression of multi-region exploration in unknown environments.

\subsection{Contribution}

Building strictly upon the theoretical foundations of standard scLTL semantics \cite{baier2008principles, kloetzer2016multi}, the original PDT framework \cite{chen2025real, chen2024fast}, and the general regret formulation \cite{blackwell1956analog, zhao2023explore, zhao2025no}, we propose a comprehensive solution framework to address the three key challenges in temporal logic task allocation for PKE, including difficult formal representation, inaccurate explicit measurement, and high computational complexity. The core contributions are summarized as follows:

We propose the Region Binding Atomic Proposition (RbAP), which explicitly binds atomic propositions to sets of regions instead of individual locations. While preserving compatibility with Deterministic Finite Automata (DFA) for task progress monitoring, it provides a rigorous formal foundation for integrating environmental uncertainty into temporal logic task modeling.

We establish a regret-based optimization framework, which resolves the inherent dilemma between exploration (gathering unknown environmental information) and exploitation (executing tasks based on known information). Introducing heuristic method to  minimize the maximum regret across all possible environmental instantiations, which ensures robust performance while balancing long-term information gain and short-term task efficiency.

We extend the PDT architecture to handle uncertain environments, proposing the E-PDT framework integrated with a regret-based BnB pruning strategy. Leveraging the min-max regret principle to prune suboptimal branches in planning stage, E-PDT effectively reduces the computational complexity induced by environmental uncertainty while maintaining near-linear scalability with respect to the number of robots and robot types.

Rigorous theoretical analysis demonstrates the feasibility and completeness of our framework. Extensive numerical experiments, simulation experiments, and physical experiments validate that our framework outperforms MILP (Mixed-Integer Linear Programming) based methods with significant advantages in both solution speed and solution quality.

\subsection{Organization}

Sec. II reviews foundational knowledge of temporal logic. Sec. III formulates the problem under investigation. Sec. IV details the E-PDT algorithmic framework. Sec. V conducts theoretical analyses of the framework’s feasibility, completeness, and computational complexity. Sec VI reports numerical, simulation, and physical experiments to validate the proposed approach. Sec VII provides discussions and conclusions.

\section{PRELIMINARIES}
Let $AP$ be a finite, non-empty set of atomic propositions. The set of LTL formulas is defined by the following grammar, using a minimal set of operators:
\begin{equation}
    \phi ::= p \mid \neg \phi \mid \phi_1 \wedge \phi_2 \mid \bigcirc \phi \mid \phi_1 \mathcal{U} \phi_2
\end{equation}
where $p \in AP$, negation $true$, $\neg$ and conjunction $\wedge$ are propositional logic operators, and $\bigcirc$ (Next) and $\mathcal{U}$ (Until) are two fundamental temporal operators. For convenience, the other propositional logic operators such as $\vee$ (disjunction), $\rightarrow$ (implication), $false$ and temporal operators such as $\lozenge$ (Finally), $\square$ (Globally) are also defined. More details about LTL syntax and semantics are referred to Principles of Model checking \cite{baier2008principles}. 

In this paper, we are working on a particular class of LTL formulae called syntactically co-safe LTL (scLTL), which means the $\neg$ (negation) can only be applied before atomic propositions and Globally ($\square$) will not appear in scLTL.

An infinite word satisfying scLTL has finite good prefix, which is a finite word defined as an finite sequence $o = o_0o_1\dots o_n \in (2^{AP})^\omega$ on the alphabet $2^{AP}$, where $o_i\in2^{AP}$ with $\forall i \in \mathbb{N}$ and $\omega$ denotes an infinite repetition. We denote the prefix of $o$ end to index $j$ as $o[:j]$, the suffix of $o$ starting at index $i$ as $o[i:]$, and the path from index $i$ to $j$ as $o[i:j]$ with $i < j$. The semantics of scLTL-formulae for finite words over alphabet $2^{AP}$ is as follows:
\begin{flalign*}
    &\ o \models true & \\
    &\ o \models p \Leftrightarrow p \in o_0 & \\
    &\ o \models \neg\phi \Leftrightarrow o \nvDash \phi & \\
    &\ o \models \phi_1 \wedge \phi_2 \Leftrightarrow o \models \phi_1, o \models \phi_2 & \\
    &\ o \models \bigcirc \phi \Leftrightarrow o[1:] \models \phi & \\
    &\ o \models \phi_1 \mathcal{U} \phi_2 \Leftrightarrow \exists i \geq 0 \; s.t. \; o_i \models \phi_2, \; \forall j \in [0, i),  o_j \models \phi_1 
\end{flalign*}

An scLTL formula can be converted to a Deterministic Finite Automaton (DFA).

\textit{Definition 1}: ({\bf Deterministic Finite Automaton}) A DFA is a 5-tuple \textit{B} = (\textit{S, $S_0$, $\Sigma$, $\delta$, $S_F$}), where $S$ denotes a finite set of states, $S_0 \in S$ denotes the set of initial states, $\Sigma = 2^{AP}$ denotes the alphabet, $\delta : S \times \Sigma \rightarrow 2^S$ denotes a transition function, and $S_F \in S$ denotes the set of accepting states.

The words which satisfy the scLTL formula can be captured by the DFA $A$. Given a word $o$, a run on it denoted by $s = s_0s_1s_2\dots s_{n+1}$ with $s_{i+1} \in \delta (s_i, o_i)$ is called accepting if $s_{n+1} \in S_F$.

\section{PROBLEM FORMULATION}
This paper investigates the problem of task allocation for a heterogeneous multi-robot system operating in a PKE, subject to complex temporal constraints specified in scLTL. The primary objective is to direct a fleet of robots to collect various resources distributed across distinct regions, in a manner that both satisfies the given scLTL specifications and operates under a regret minimization framework. The conventional method of defining atomic propositions by associating each with an individual region is insufficient for our problem formulation. Therefore, we propose RbAP, where atomic propositions are defined over sets of regions.  In what follows, we provide a detailed description of the PKE, RbAP, and the formulation of regret within our model.

\subsection{Partially Known Environments}
To ensure rigor, the PKE in our framework is formally defined by the following set of assumptions \cite{zhao2025no}. Through this abstraction method, we can effectively formalize practical engineering applications such as Urban Search and Rescue (USAR), which typically feature known topological structures but possess dynamic semantic uncertainties.

\begin{itemize}
\item {\bf Known Environmental Structure and Potential Resources}: The topology of the environment is considered known, and the set of possible resource types it may contain is also known beforehand for each region.
\item {\bf Guaranteed Resource Anchor}: 
For each type of resource, the robots are given knowledge of exactly one designated region where its presence is guaranteed. The existence of this resource in all other potential regions is initially uncertain and must be discovered.
\item {\bf Local Observability and Static Conditions}: 
Upon a robot's arrival at a region, the true presence or absence of resources in that location is unambiguously and permanently revealed to it, and this state of resource distribution is fixed. 
\end{itemize}

To formally construct our model for planning under uncertainty, we now introduce a hierarchy of definitions, starting with the workspace and culminating in the comprehensive model of the PKE.

\textit{Definition 2}: ({\bf Workspace and Region Belief}) The workspace is a finite set of regions $R = \{r_1, r_2, \dots. r_n\}$. A single region belief maintained by the HMRS, is denoted by $r_{b,i} = (r_i, pot_i, cer_i)$, where $pot_i$ is the set of resources that may exist, and $cer_i$ is the set of resources that are certain to exist.

The collection of all individual region beliefs forms the robot's overall knowledge of the world belief, which we distinguish from the actual, true state of the environment.

\textit{Definition 3}: ({\bf World Belief}) A world belief $W_b$ is the set of all region beliefs, representing the robot's current knowledge, denoted by $W_b = \{r_{b,i} \vert r_i \in R\}$. The robot's initial knowledge is the initial world belief, $W_{b0}$. 

\textit{Definition 4}: ({\bf Environmental Instantiation}) An environmental instantiation $E$ represents a determined instance of the world that has the same structure as a world belief, where each region is denoted by $r_{e,i} = (r_i, pot_i, cer_i)$ and $pot_i = \emptyset$. 

To mathematically connect the robot's subjective belief with this objective reality, we introduce the crucial notion of compatibility.

\textit{Definition 5}: ({\bf Compatibility}) An environmental instantiation $E$ is said to be compatible with a world belief $W_b$, denoted by $E \in W_b$, if
$$
cer_{e,i} \subseteq (cer_{b,i} \cup pot_{b,i})
$$
for every environmental instantiation $r_{e,i} \in E$ and corresponding region belief $r_{b,i} \in W_b$.

Equipped with these foundational definitions, we can now precisely define the overarching Partially Known Environment as the set of all possible worlds consistent with the robot's initial knowledge.

\textit{Definition 6}: ({\bf Partially Known Environment}) The Partially Known Environment $\mathcal{E}$ is the set of all environmental instantiations $E$ that are compatible with the robot's initial world belief $W_{b0}$. The set of possible environments derived from any belief $W_b$ is similarly denoted as $\mathcal{E}_{W_b} = \{E|E \in W_{b}\}$.

\subsection{Heterogeneous Multi-Robot Systems}

We consider an HMRS tasked with the mission. The system consists of a set of $n_a$ robots, denoted $A = \{ a_1, a_2, \dots, a_{n_a}\}$, which are classified into $n_t$ distinct types from the set $T = \{ t_1, t_2, \dots, t_{n_t} \}$. We define that each robot belongs to exactly one type, $a_i^j$ means robot $i$ is of type $j$. In our formulation, the heterogeneity among robots is manifested through their resource discovery and carry capabilities. We define that $T_j$ contains all robots of type $j$, then we have $n_a = \Sigma_j Card(T_j)$ and $T_j \cap T_k = \emptyset$, $\forall j \neq k$.

\subsection{Regions Binding Atomic Proposition}

\textit{Definition 7}: ({\bf Regions Binding Atomic Proposition}) An RbAP is a 5-tuple $ap = (l_p, m_p, R_p, GR_p, A_p)$, where $l_p$ denotes the label, $m_p$ denotes the corresponding resources set, $R_p$ denotes the corresponding regions, $GR_p$ denotes the region that guaranteed to have $m_p$, and $A_p$ denotes the robot set required for the task.

\textit{Example 1}: Let us consider a resource collection scenario to instantiate the problem formally. We define a mission environment $R = \{r_i\},0\leq i\leq 8$, where a team of robots, $A = \{a_i\},1\leq i\leq 4$, is initially co-located in a base region, called $r_0$. The mission requires the collection of two distinct resource types, denoted $\alpha$ and $\beta$. $a_1$ and $a_2$ are of type 1 for resource $\alpha$, while $a_3$ and $a_4$ are of type 2 for resource $\beta$. Regarding the distribution, resource $\alpha$ may be present in the set of regions $\{r_1, r_2, r_3, r_4\}$. Within this set, regions $\{r_1, r_2, r_3\}$ are potential locations where the presence of $\alpha$ is uncertain. Region $r_4$ serves as the guaranteed region, where the presence of $\alpha$ is known a priori. Resource $\beta$ is distributed across regions $\{r_5, r_6, r_7, r_8\}$. Its presence is uncertain in potential regions $\{r_5, r_6, r_7\}$, while it is guaranteed to exist in region $r_8$. To formalize the mission objectives, we associate these collection tasks with two RbAP, $ap_1$ and $ap_2$ respectively. As for $ap_1$, $m_p=\alpha$, $R_p = \{r_1, r_2, r_3, r_4\}$, $GR_p = r_4$, $A_p = \{a_1, a_2\}$,and the atomic proposition $ap_2$ is defined analogously. The overall mission is to eventually acquire both resources without strict ordering, which is concisely captured by the scLTL formula $\phi = (\lozenge ap1) \wedge (\lozenge ap2)$.

\subsection{Policy and Regret}

The classic exploration-exploitation dichotomy is insufficient to characterize a policy operating under the constraint of temporal logic, since decision-making must be augmented by the progress of the logical task specification. We therefore propose the following formulation:

\textit{Definition 8}: ({\bf Policy}) A policy is a 2-tuple $\pi = (\rho, \xi)$, where $\rho$ denotes the sub-task derived from the decomposition of the automaton, $\xi$ denotes the choice set for each proposition between exploration and exploitation.

Within our framework, the decision-making process implements a hybrid concurrency strategy. To maximize system concurrency, the planner prioritizes the transition edge comprising the maximal set of atomic propositions at each decision step. Subsequently, by defining action modes at the level of these individual atomic propositions, the system can simultaneously pursue exploitation for certain objectives while conducting exploration for others within the same planning step. This process yields a policy sequence, denoted by $\pi = \pi_1\pi_2\dots\pi_n$. Specifically, the strategy prioritizes dispatching robots to guaranteed regions to secure mission objectives, while any surplus robots are allocated to potential regions to maximize information gain. Following the procedure described above, this policy sequence is then expanded into a \textit{plan}, $\tau = \tau_1\tau_2\dots\tau_n$, which explicitly specifies the sequence of regions to be visited by each individual robot. This fine-grained control allows for the parallel execution of diverse task components, thereby maximizing the utilization of heterogeneous robotic resources and accelerating the overall mission completion.

As for the regret, a conventional approach for evaluating task allocation is to minimize the worst-case cost. This method, however, struggles to account for the long-term benefits of exploration, thus failing to resolve the classic exploration-exploitation dilemma. To address this limitation, we introduce the concept of regret \cite{blackwell1956analog} to balance the competing objectives of exploring uncertain regions and exploiting known information. Regret is formally defined as the difference between the actual cost accumulated by the decisions made thus far, and the cost of the optimal policy that could have been achieved in hindsight. 

\textit{Definition 9}: ({\bf Regret}) Given the world belief $W_{b}$, the policy set $\Pi(\mathcal{E}_{W_b}, \phi)$, and a task formulated by an scLTL $\phi$, the regret of policy $\pi$ is defined by
\begin{equation}
\label{regret}
    reg_{W_b}(\pi) = \max \limits_{E\in\mathcal{E}_{W_b}} [cost(\pi,E)-\min\limits_{\pi'\in\Pi} cost(\pi', E)]
\end{equation}

In (\ref{regret}), $cost(\pi, E)$ means the cost of applying policy $\pi$ to environmental instantiation $E$. The latter term, which we will refer to as the hindsight-optimal cost, is the minimum achievable cost for $E$ and is found by selecting the best policy $\pi'$ from the set of all valid policies $\Pi(\mathcal{E}, \phi)$. The subtraction of these two costs yields the regret for a single environment instantiation. Finally, the $\max$ operator defines the overall regret of policy $\pi$ as its worst-case performance across world belief. This metric quantifies the opportunity cost incurred by the offline contingent planning due to incomplete information and the necessity of information gathering. By minimizing regret, the robot team can effectively navigate the trade-offs inherent in PKE.

\begin{figure}[tb]
      \centering
      \includegraphics[width=0.25\textwidth]{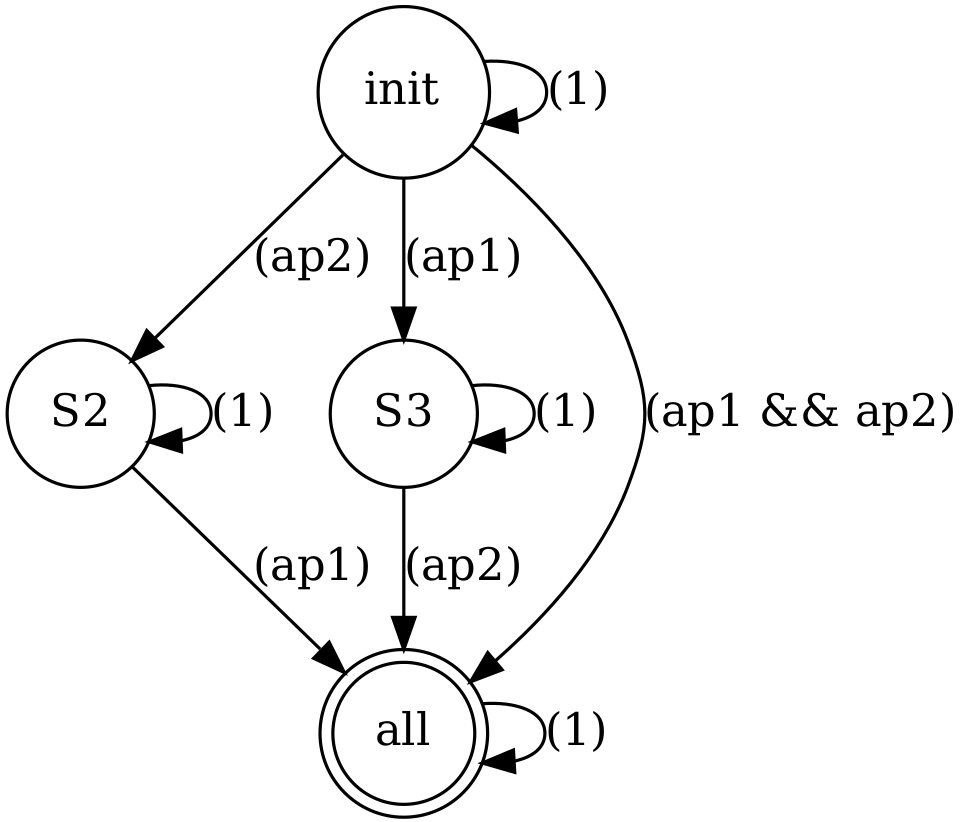}
      \caption{Automaton figure of example scLTL $\phi = ap_1 \wedge ap_2$.}
      \label{example}
\end{figure}

\textit{Example 2}: ({\bf Example 1 Cont.}) As illustrated in Fig. \ref{example}, there are three sub-task options available from the initial state; we choose the transition with most atomic propositions $ap_1 \wedge ap_2$. Assuming both $ap_1$ and $ap_2$ have available guaranteed regions and potential regions, the algorithm generates $2^2 = 4$ distinct hybrid policies: $\pi_1=$ (Exploit $ap_1$, Exploit $ap_2$), $\pi_2=$ (Exploit $ap_1$, Explore $ap_2$), $\pi_3=$ (Explore $ap_1$, Exploit $ap_2$), and $\pi_4=$ (Explore $ap_1$, Explore $ap_2$). Consider the hybrid policy $\pi_2$ as an instance. For $ap_1$, the planner allocates robot $a_1$ to the guaranteed region $r_4$ while simultaneously dispatching the surplus robot $a_2$ to explore the potential region $r_1$. Concurrently, for $ap_2$, robots are assigned to explore potential regions $\{r_5, r_6\}$. This policy containing 3 potential regions, which leads to $2^3 = 8$ distinct child nodes. Assuming the maximum regret among these eight scenarios is 14, this value is assigned to $\pi_2$.

Finally, we formally formulate the problem that we solve in this paper as follows:

\textit{Problem 1: Given the initial PKE $E$, the heterogeneous multi-robot system $A$, and an scLTL task $\phi$ represented by RbAP, the goal is to obtain a policy $\pi^* = \pi_1\pi_2\dots\pi_n$ and the corresponding plan $\tau^* = \tau_1\tau_2\dots\tau_n$ satisfiy the task $\phi$ such that $reg_\mathcal{E}(\pi^*) \leq reg_\mathcal{E}(\pi'), \forall \pi' \in \Pi(\mathcal{E},\phi)$.}

\section{TASK ALLOCATION AND PLANNING}

\begin{figure}[tb]
      \centering
      \includegraphics[width=0.48\textwidth]{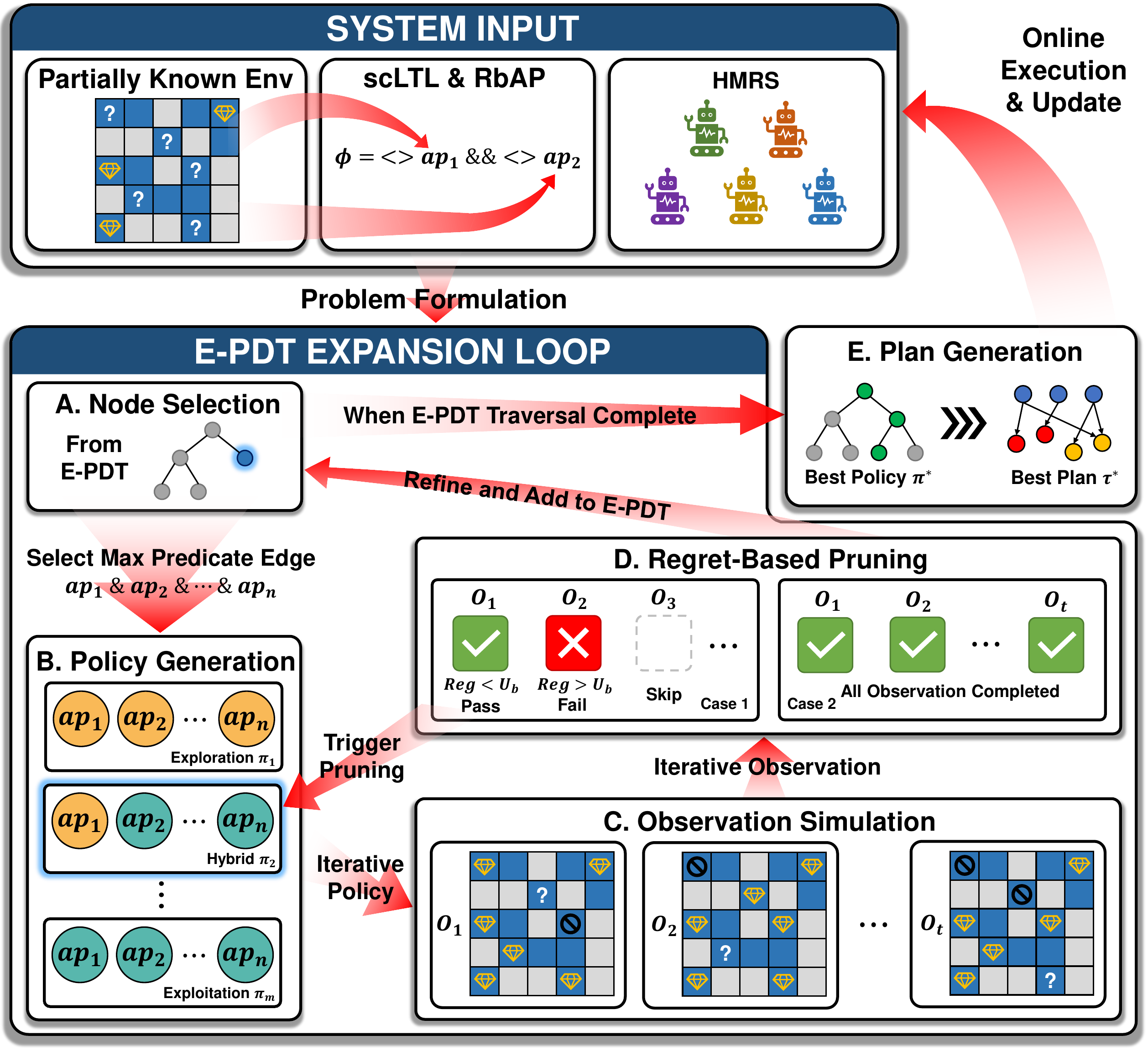}
      \caption{Framework of E-PDT. The framework takes the PKE, scLTL specifications, and HMRS as inputs. The core mechanism is the E-PDT Expansion Loop, acting as offline contingent planning, which iteratively processes tasks through five stages: (A) Node Selection, (B) Policy Generation, (C) Observation Simulation, (D) Regret-Based Pruning to filter suboptimal branches, and (E) Plan Generation. The final output is the optimal policy $\pi^*$ and plan $\tau^*$ for  online branches selection.}
      \label{zongtu}
\end{figure}

To address the decision-making challenges under uncertainty defined in Problem 1, we propose a regret-minimizing planning framework, as shown in Fig. \ref{zongtu}. This approach builds upon the PDT structure \cite{chen2024fast, chen2025real}, which was originally designed for scalable planning in fully known environments. However, because accurate prior probabilities of resource distribution are typically unavailable in PKE, our framework fundamentally diverges by handling uncertainty non-stochastically. We introduce two key modifications to explicitly reason about incomplete information for robust decision-making. First, semantic uncertainty, encapsulated within the RbAP model, strictly dictates the tree expansion. We explicitly consider a set of distinct policies as defined in \textit{Definition 8}, anticipating all possible stochastic outcomes of an exploratory action to generate corresponding observation branches, which yield different $Cost$ and $Reg$ values for the resulting child nodes. Second, we integrate a min-max regret pruning mechanism directly into the expansion loop. Rather than navigating uncertainty by calculating expected probabilities, the framework evaluates the Min-Max Regret across all branches. By computing the regret value for each potential outcome sequentially and discarding suboptimal policies on-the-fly, this mechanism minimizes computational overhead while robustly bounding the opportunity cost of acting under incomplete information. The specific algorithm details are as follows.

\subsection{Extended Planning Decision Tree}

\begin{algorithm}[tb] 
    \caption{Extended Planning Decision Tree}
    \label{alg::E-PDT}
    \SetAlgoLined 
    \KwIn{scLTL $\phi$, Partially Known Environment $E$, Multi-Agent System$A$}
    \KwOut{Plan $\tau^*$}
    Initialize E-PDT $T$\\
    \While{$\exists Nd \in T$, s.t. $Nd.Trav == True$}{
        $Nd\_to\_expand = T$\\
        \For{$Nd \in Nd\_to\_expand$}{
            \If{$Nd.Trav == True$}{
                \Continue\\
            }
            $Nd_{sub} = \textsf{treeExpansion}(Nd)$\\
            $T.\textsf{addNode}(Nd_{sub})$\\
            $Nd.Trav = True$\\
        }
    }
    $\tau^* = \textsf{getPlan}(T)$\\
    \KwRet{$\tau^*$}\\
\end{algorithm}

\textit{Definition 10}: ({\bf Extended Planning Decision Tree}) The E-PDT $T_D$ is built on the basis of a set of nodes $Nd_i, i \in \mathbb{N}$, which is formally defined as a tuple
$$
Nd_i = (Pol, A_s, EP_{pre}, ET_{pre}, Cost, B_s, Reg, O)
$$

where
\begin{itemize}
    \item $Pol$ is the policy $\pi_j$, as defined in \textit{Definition 8}.
    \item $A_s$ records which specific RbAP each individual robot is assigned to within a hybrid policy.
    \item {$ET_{pre}$, $EP_{pre}$, $B_s$} are inherited from the original PDT, $ET_{pre}$ is predicted robot locations, $EP_{pre}$ is predicted robot time stamp completing the subtask corresponding to RbAP, and $B_s$ is the DFA state.
    \item $Cost$ is detailed below.
    \item $Reg$ is the regret of policy $\pi_j$, as defined in \textit{Definition 9}
    \item $O$ is a world belief defined in \textit{Definition 3}.
\end{itemize}

To facilitate exposition and maintain consistency with the foundational PDT framework \cite{chen2025real}, we inherit several basic node attributes, namely $Parent(Nd_i)$ to denote the parent node, $Prog = \{pre, acc\}$ for the mission stage, $PreS(Nd_i)$ for the set of previously visited automaton states within the same stage, and $Trav = \{true, false\}$ as a traversal flag. Specifically, within the mission stage $Prog$, $pre$ (prefix) signifies any intermediate state of the automaton prior to the final completion of the task, corresponding to the ongoing execution phases where robots are actively resolving sub-tasks but the overall high-level specification has not yet been fully satisfied. Conversely, $acc$ (accept) denotes the accepting state of the automaton, indicating the termination of the mission where all required sub-tasks have been successfully executed. While we retain these structural attributes to model the task progression, a fundamental departure in our framework lies in the evaluation of the node $Cost$, which is a critical component for decision-making under uncertainty. Whereas the original PDT defines cost as the deterministic task makespan $\max(ET_{pre})$, our approach considers the stochastic outcomes of exploration. Assuming unconstrained inter-robot communication, $Cost$ is defined based on the observation result. $Cost$ is the completion time of the first successful robot if at least one robot finds the resource. Otherwise, $Cost$ is the maximum time taken by any dispatched robot. Moreover, we define the crucial items $Pol$, $Reg$ and $O$ to serve for PKE. 

With the node attributes formally defined, we now introduce the overarching framework of the proposed E-PDT algorithm in Algorithm \ref{alg::E-PDT}. The search process initiates from the root node and iteratively expands the tree until no further nodes can be generated, ultimately yielding the executable policies for the multi-robot fleet. The algorithmic core of this framework lies in the \textsf{treeExpansion} function, as shown in Algorithm \ref{alg:tree_expansion}, which explicitly dictates how each parent node is systematically expanded to generate child nodes under environmental uncertainty. Furthermore, upon the termination of the tree expansion, the \textsf{getPlan} function is utilized to extract a robust contingent policy tree featuring dynamically switchable execution branches. The specific mechanisms underlying both the node expansion process and the final plan generation will be thoroughly elaborated in the subsequent subsections.

\subsection{Policy-based Tree Expansion}
    
    



\begin{algorithm}[tb]
    \caption{Policy-Based Tree Expansion}
    \label{alg:tree_expansion}
    \SetAlgoLined
    \KwIn{Parent Node $Nd$}
    \KwOut{A list of child nodes $Nd_{sub}$}
    
    $Nd_{sub} = \emptyset$\\
    Initialize the $subtasks$\\
    
    \For{$st$ $\in$ $subtasks$}{
        $Pols = \textsf{generatePolicy}(st)$\\
        
        \For{$\pi \in Pols$}{
            \For{$ap \in \pi.\rho$}{
                \uIf{$\pi.\xi == Exploitation$}{
                    $\textsf{alloExploit}(ap)$\\
                }
                \uElseIf{$\pi.\xi == Exploration$}{
                    $\textsf{alloExplore}(ap)$\\
                }    
                \textbf{end}
            }
            $obs = \textsf{geneObs}()$\\
            \For{$o \in obs$}{
                $\textsf{modifyAutomatonState}(st,o)$\\
                $ChildNd = \textsf{geneChildNode}(\pi, o)$\\
                $\textsf{computeRegret}(Nd_{sub})$\\
                $\textsf{pruneByRegret}(Nd_{sub})$\\
                $Nd_{sub}.\textsf{add}(ChildNd)$\\
            }
        }
    }
    \KwRet{$Nd_{sub}$}\\
\end{algorithm}

The computational efficiency of the PDT for rapid task allocation stems from its strategic decision to forgo detailed path planning for individual robots. Specifically, it leverages sub-tasks decomposed from a task automaton to guide a greedy, single-step optimal robot assignment process, resulting in an algorithmic complexity that scales linearly with the number of robots. Our work extends this framework to address decision-making under uncertainty. 

Algorithm \ref{alg:tree_expansion} outlines the procedure for expanding a single parent node $Nd$. In the initialization phase (line 2), the algorithm selects the transition edge associated with the maximal set of atomic propositions derivable from the current automaton state. By prioritizing system concurrency, this greedy heuristic aims to fully utilize the heterogeneous fleet. While such high-concurrency exploratory actions inherently carry higher theoretical regret bounds, due to the cumulative risk of simultaneous exploration failures, they are essential in partially known environments. Specifically, they maximize concurrent information gain, deliberately trading short-term regret bounds for long-term robustness and reduced actual execution costs. Subsequently, the \textsf{generatePolicy} function (line 4) processes this selected subtask by combinatorially assigning exploration or exploitation modes to each atomic proposition to form hybrid policies.

After generating the hybrid policy, the concrete robot allocation is executed through two distinct functions in Algorithm \ref{alg:tree_expansion}: \textsf{alloExploit} and \textsf{alloExplore}. Specifically, the \textsf{alloExploit(ap)} function inherently executes a two-step allocation. It first strictly assigns a robot team to the guaranteed region to secure the task objective, and subsequently directs any remaining surplus robots to explore the associated potential regions in parallel in order to gain more information. Conversely, the \textsf{alloExplore(ap)} function allocates robots exclusively to potential regions for pure information gathering.

To solve the optimal robot-to-region assignments within these two functions, an underlying two-level greedy mechanism is utilized. At the higher level, an inter-region assignment is performed through an iterative and competitive greedy process. In each step, it evaluates all currently unassigned regions to determine the minimum potential cost for visiting each one with the available robots. After that, it greedily selects and commits to the single best region-team pairing that yields the lowest makespan. The robots in this committed team are subsequently removed from the pool of available resources, and the algorithm proceeds to the next decision step, repeating the evaluation and selection process with the remaining regions and robots until the allocation is complete. At the lower level, this process is supported by an intra-team composition mechanism that calculates the cost for any single region. Specifically, the predicted arrival time for an individual robot assigned to a target region is calculated as the sum of its accumulated execution time from previously assigned tasks and the travel time for the current task, that is the path distance to the target region divided by the robot's moving velocity. To evaluate the cost, the mechanism greedily selects a coalition of available robots that satisfies the task's requirements with the minimum possible makespan, defined by the arrival time of the last robot in the team. This cost serves as the primary metric for the high-level iterative selection process.

Once a policy is generated and the corresponding robots are notionally dispatched, the framework must reason about the uncertain outcomes of exploration. This is handled by the function \textsf{geneObs} (line 13), which projects a set of all possible environmental observations. For instance, if two potential regions are explored, \textsf{geneObs} would generate four potential outcomes: success in both, either one, or neither. Therefore, four child nodes with same policy and plan but different observations are generated, which are used to update the state of DFA.

A critical challenge remaining before a child node can be finalized is determining its resultant automaton state. Unlike methods designed for fully known environments, our framework operates under uncertainty where the outcome of an action is not known a priori. The translation of an observation into formal task progress, function \textsf{modifyAutomatonState} (line 15), is essential because the outcomes of an exploration policy are inherently stochastic; a single observation could satisfy no propositions, one, or even multiple propositions simultaneously. Function \textsf{modifyAutomatonState} first identifies the fulfilled RbAP by a given observation, and then consults the automaton graph to find all transitions from the current state that are enabled by this set. Crucially, to resolve ambiguity when multiple paths forward are possible, the procedure selects the maximal transition. In contrast, for an exploitation policy where success is assured, this complex evaluation is unnecessary, and the automaton state advances deterministically. 

The final step in the expansion process involves instantiating a child node for each potential observation $O$ and populating the E-PDT tuple fields (Definition 10) conducted by \textsf{geneChildNode}. Crucially, this generation phase is tightly integrated with the Branch-and-Bound (BnB) strategy to ensure computational efficiency, conducted by \textsf{computeRegret} and \textsf{pruneByRegret}. The instantiation of the node tuple is explicitly defined as follows. First, the parent node first generates the candidate policy $Pol$. Subsequently, the greedy algorithm determines the specific task allocation $A_s$ for each robot at this node, their corresponding projected locations $EP_{pre}$, the required execution times $ET_{pre}$, and the resulting $Cost$. By simulating potential environmental outcomes, a distinct child node is generated for each possible observation $O$ under this policy, and the actual task automaton state $B_s$ is evaluated under different observations. Finally, the regret value $Reg$ is computed based on this synthesized information. Instead of blindly computing all nodes, the algorithm continuously performs pruning checks. If a node's computed regret violates the global upper bound, the expansion for the corresponding policy is aborted. That is, the regret computation for all sibling nodes under this policy is terminated, and these nodes are permanently discarded from further consideration in subsequent expansions. Comprehensive details on these regret calculating and pruning mechanisms are presented in the following sections.

\subsection{Regret Computation}

While \textit{Definition 9} provides a formal and complete description of regret, its direct application within a fast planning algorithm is computationally intractable. Specifically, calculating the hindsight-optimal cost term, $\min_{\pi'\in\Pi(\mathcal{E},\phi)}cost(\pi',E)$, would require solving the entire planning problem for every possible instantiation of the environment $E\in\mathcal{E}$. To make the regret calculation tractable during the tree expansion, we therefore introduce a heuristic-based method to approximate (\ref{regret}) for node $Nd$ as function \textsf{computeRegret} in Algorithm \ref{alg:tree_expansion}.
\begin{equation}
    \label{approximate reg}
    \tilde{reg}(\pi) = \max_{Nd\in Nds(\pi)}[Nd.Cost + H(Nd.O) - opt(Nd.O)]
\end{equation}
where $Nds(\pi)$ means the node set by the adoption of policy $\pi$. 

In (\ref{approximate reg}), $Nd.Cost$ represents the accumulated cost from the root, $H(Nd.O)$ estimates the future cost-to-go to complete the remaining tasks, and $opt(Nd.O)$ represents the minimum possible cost to satisfy the specification from initial given the current belief. Crucially, to ensure the theoretical validity of the pruning, we introduce the optimistic resource assumption. We define $W_{opt}$ as an idealized environment where every potential region in the current belief $Nd.O$ is assumed to validly contain the resource. Conversely, $W_{real}$ corresponds to the ground truth where potential regions may be empty. Both $H(Nd.O)$ and $opt(Nd.O)$ are calculated strictly within $W_{opt}$, assuming the agent is simply directed to the nearest region containing the resource, which means that (\ref{approximate reg}) essentially computes the regret within the optimistic world. As we will formally prove in \textit{Lemma 1} in Section V-A, this formulation guarantees that the estimated regret $\tilde{reg}(\pi)$ is a strict lower bound of the true regret in $W_{real}$, ensuring the admissibility of our BnB strategy.

\subsection{Regret-Based Branch-and-Bound Pruning}

To mitigate the combinatorial explosion caused by the branching of environmental observations, we integrate a BnB strategy into the tree expansion. Our approach guaranties the retrieval of the min-max regret solution by maintaining a global upper bound, denoted $U_{best}$, which represents the minimum max-regret value among all complete plans discovered so far. The pruning mechanism operates on the principle that the maximum regret of a policy is determined by its worst-case observation outcome. Therefore, if any partially evaluated outcome of a candidate policy exhibits a regret exceeding $U_{best}$, the policy is provably suboptimal. This allows for an early-exit strategy, formally defined by the following traversal rules:

\textit{Definition 11}: ({\bf Pruning Rules}) The expansion and pruning of the E-PDT are governed by the following rules:

\begin{enumerate}
    \item A node is pruned if its mission stage indicates that the automaton has reached the accepting state. In the context of scLTL, this signifies that a finite good prefix has been successfully generated, meaning the overall finite-horizon task is fully accomplished and no infinite suffix stage is required.
    \item A generated node is pruned if both its automaton state and observation outcome have already been encountered with a smaller regret value within the current mission stage.
    \item For a candidate policy $\pi$, its potential environmental observations $\{o_1, o_2, \dots, o_m\}$ are evaluated sequentially. During this process, if the regret of any single observation instance $o_k$, denoted as $r\tilde{e}g(\pi, o_k)$, satisfies $r\tilde{e}g(\pi, o_k) \ge U_{best}$, the evaluation of policy $\pi$ is immediately terminated. The policy is deemed suboptimal, and all its associated nodes ($\{o_1, o_2, \dots, o_m\}$) are pruned.   
\end{enumerate}

Rules 1 and 2 ensure the termination and acyclicity of the search. Rule 3 constitutes the algorithmic core of our efficiency improvement, denoted by function \textsf{pruneByRegret} (line 18) in Algorithm \ref{alg:tree_expansion}. Unlike naive min-max approaches that must compute the regret for all possible observations to determine a policy's worst-case performance, our strategy employs an early-exit mechanism. Since the overall regret of a policy is determined by its maximum component (Eq. 3), finding just one outcome that performs worse than the current global best ($U_{best}$) is sufficient to prove the policy's suboptimality. This allows the algorithm to discard the policy without expending computational resources on the remaining, unexamined observation branches.

The significance of this strategy is two-fold. Computationally, the sequential evaluation mechanism enables aggressive pruning: a policy is discarded immediately upon detecting a single outcome that exceeds the upper bound, thereby avoiding the evaluation of remaining observation branches. Qualitatively, it ensures robustness, guiding the framework to identify strategies that remain effective even under the most unfavorable environmental conditions.

\textit{Example 3}: ({\bf Example 2 Cont.}) Suppose the current global best regret is $U_{best} = \tilde{reg}(\pi_2) = 10$. The algorithm begins evaluating a new policy $\pi_{new}$, which has three potential observation outcomes.The algorithm computes the regret for the first observation, yielding 5. Since $5 < 10$, the evaluation continues.The second observation yields a regret of 12. Since $12 \ge 10$, Rule 3 is triggered, and the algorithm immediately stops processing $\pi_{new}$. The third observation outcome is never computed, and $\pi_{new}$ is pruned from the tree. 

\subsection{Plan Generation}

The final output of our framework, generated by the function \textsf{getPlan} (Algorithm \ref{alg::E-PDT}, line 13), is not a single linear plan but a robust contingent policy tree. This tree represents the complete set of optimal policies and the corresponding concrete robot assignments for every reachable state of belief. The optimal policy sequence, $\pi^*$, is obtained directly from the selection and pruning process detailed in the previous section. The corresponding executable plan, $\tau^*$, which details the target regions for each robot, is then derived from the nodes associated with this optimal policy by extracting the task allocation data. During the subsequent online execution phase, the robot team executes the generated contingent plan $\tau^*$. The real-world observation resulting from this action then determines the subsequent path within the plan, dictating the next set of robot assignments to be executed.

\textit{Example 4}: ({\bf Example 3 Cont.}) Let us assume the contingent policy tree has been generated, and the online execution phase starts. The robot team begins at the root node and executes the initial optimal plan corresponding to the optimal policy $\pi^*$. While the offline planning phase accounted for this policy's four potential outcomes by creating four distinct child nodes, the single, real-world observation made upon execution deterministically selects which of these precomputed branches to follow. The process is then repeated from this new node, continuing until a terminal node representing task completion is reached.

\section{ALGORITHM ANALYSIS}

In this section, we first establish the feasibility of the generated plans, ensuring they satisfy the scLTL specifications. Subsequently, we prove the completeness of the algorithm. Next, we demonstrate that the proposed regret-based BnB strategy is admissible by proving that the heuristic estimated regret serves as a strict lower bound for the true regret. Finally, we derive the computational complexity of the algorithm, analyzing how the framework manages the exponential search space introduced by environmental uncertainty.

\subsection{The Performance of E-PDT}

\textit{Theorem 1}: State that any optimal plan $\tau^*$ generated by E-PDT is feasible with the scLTL formula $\phi$, if there exists a plan satisfying $\phi$.

\begin{proof}
The feasibility of the generated plan $\tau^*$ is guaranteed by the construction of the E-PDT. The plan is extracted from a path of nodes $\mathcal{P} = (Nd_0, \dots, Nd_n)$ in the tree. According to the \textsf{treeExpansion} function, the transition between any two consecutive nodes, $Nd_{i-1}$ and $Nd_i$, corresponds to a valid transition in the DFA, i.e., $Nd_i.B_s \in \delta(Nd_{i-1}.B_s, o_{i-1})$. Furthermore, the algorithm extracts a plan only from a path that terminates at a node $Nd_n$ where $Nd_n.Prog = acc$, which means its automaton state $Nd_n.B_s$ is an accepting state. Therefore, the plan corresponds to an accepting run on the DFA and is guaranteed to be feasible.
\end{proof}

\textit{Theorem 2}: If a feasible plan satisfying the scLTL formula $\phi$ exists, E-PDT is guaranteed to find one if the min-max pruning rule (Rule 3 in Definition 11) is disabled.

\begin{proof}
The proof relies on the exhaustive nature of the \textsf{treeExpansion} function when regret-based pruning is disabled. Let us assume a feasible plan $\tau$ exists, which corresponds to a valid execution trace $tr(\tau) = (\pi_0, O_0, \pi_1, O_1, \dots, \pi_{n-1}, O_{n-1})$. At any given node, the \textsf{treeExpansion} function systematically generates child nodes by considering all valid subtasks from the DFA, creating all applicable policies for each subtask, and branching on every possible environmental observation from an exploration policy. By this exhaustive construction, the un-pruned E-PDT is the complete tree of all possible valid execution traces. Since $tr(\tau)$ is a valid trace, the corresponding path of nodes $\mathcal{P} = (Nd_0, Nd_1, \dots, Nd_n)$ is guaranteed to be constructed within this tree, and the algorithm is therefore ensured to find it.
\end{proof}

\textit{Lemma 1}: The estimated regret $\tilde{reg}(\pi)$ defined in (\ref{approximate reg}) is a valid lower bound of the true regret defined in (\ref{regret}). Consequently, the pruning strategy preserves the completeness of the algorithm.

\begin{proof}
For any policy $\pi$, its exact regret under $W_{opt}$ is
$$
reg_{W_{opt}}(\pi) = Cost(\pi, W_{opt}) - \min\limits_{\pi'\in\Pi} Cost( \pi', W_{opt})
$$
where $W_{opt} \in \mathcal{E}_{W_b}$ is a valid environmental instantiation according to Section IV-C.

For each node $Nd$ along the policy, $Nd.Cost$ is the accumulated deterministic cost from the root. The heuristic $H(Nd.O)$ underestimates the remaining cost-to-go under the optimistic assumption, yielding
$$
Nd.Cost + H(Nd.O) \leq Cost(\pi, W_{opt})
$$

Since $opt(Nd.O)$ corresponds to the optimal cost \(\min_{\pi' \in \Pi} \text{Cost}(\pi', W_{\text{opt}})\) in the optimistic world, yielding
$$
Nd.Cost + H(Nd.O) - opt(Nd.O) \leq reg_{W_{opt}}(\pi)
$$
for all $Nd \in Nds(\pi)$. Taking the maximum over all associated nodes yields
$$
\tilde{reg}(\pi) \leq reg_{W_{opt}}(\pi)
$$

By Definition 9, $reg_{W_b}(\pi)$ is the maximum regret over all compatible environmental instantiations. Since $W_{\text{opt}} \in \mathcal{E}_{W_b}$, we have 
$$
reg_{W_{opt}}(\pi) \leq reg_{W_b}(\pi)
$$

Combining the inequalities yields $\tilde{reg}(\pi) \leq reg_{W_b}(\pi)$. 
\end{proof}

\subsection{The Complexity of E-PDT}

We formally analyze the computational complexity of the proposed E-PDT algorithm, denoted as $C_{total}$, is modeled as the product of the number of nodes visited in the search tree and the computational cost required to process a single node:
$$C_{total} = C_{node} \times N_{visited}$$

Expanding a node involves generating a set of candidate hybrid policies, generated $2^{|\mathcal{AP}_{sub}|}$ policies, where $|\mathcal{AP}_{sub}|$ is the number of atomic propositions involved in the current subtask. For each policy, the dominant computational burden is the heuristic regret estimation, which requires estimating the cost-to-go for the remaining $|\mathcal{AP}|$ unfulfilled tasks. As described in Algorithm \ref{alg:tree_expansion}, the core allocation mechanism employs a greedy strategy that sorts the $n_a$ robots of $n_t$ types by estimated arrival times, with a complexity of $\mathcal{O}(n_a \log n_a + n_t)$. Since the heuristic sequentially applies this allocation logic to the remaining atomic propositions, and each subtask involves at most $|R_{s}|$ candidate regions, the total cost to process a single node is:
$$C_{node} = \mathcal{O}\left(2^{|\mathcal{AP}_{sub}|} \cdot |\mathcal{AP}| \cdot |R_{s}| \cdot (n_a \log n_a + n_t) \right)$$

The term $N_{visited}$ is fundamentally determined by the tree depth and the branching factor. In original PDT and fully known environments, the maximum search depth is inherently bounded by the task automaton size, i.e., $\mathcal{O}(|S|)$. However, in PKE, an exploration action may update the environment belief without triggering an automaton transition, such as revealing empty regions. To rigorously capture this, we account for the concurrent exploration capacity of the heterogeneous fleet. For a subtask requiring $r_j$ robots of type $j$ from an available pool of $Card(T_j)$, the fleet can simultaneously evaluate $N_{conc} = \lfloor Card(T_j) / r_j \rfloor$ regions. By incorporating this capacity, the theoretical maximum search depth $D_{max}$ is formulated as:

$$D_{max} = \mathcal{O}\left(|S| + \frac{|R_{unc}|}{N_{conc}}\right)$$

Given that a candidate policy in policy set $\Pi$ evaluating $|R_{unc}|$ potential regions generates up to $2^{|R_{unc}|}$ observation branches, the worst-case search space size becomes:
$$N_{visited}^{worst} = \mathcal{O}\left(|\Pi| \cdot 2^{|R_{unc}| \cdot {D_{max}}}\right)$$

It is important to address the exponential term $\mathcal{O}(2^{|R_{unc}|\cdot {D_{max}}})$, which is an inherent attribute of any algorithm seeking a min-max regret solution in PKE, as it necessitates exhaustively evaluating the risk of all potential stochastic outcomes arising from exploratory actions to guarantee worst-case robustness. However, despite this theoretical bottleneck, the analysis confirms that the framework inherits the excellent scalability of the original PDT architecture: the single-node computation maintains near-linear scalability $\mathcal{O}(n_a \log n_a)$ with respect to the number of robots $n_a$, and linear scalability with the number of robot types $n_t$. To effectively combat the exponential complexity introduced by environmental uncertainty, the proposed regret-based BnB strategy is integrated directly into the expansion loop. By dynamically evaluating and pruning sub-optimal branches early, this mechanism prevents the exhaustive traversal of all observation outcomes in practice. This demonstrates that the E-PDT framework, despite operating in the significantly more complex domain of PKE, retains the fundamental advantage of scalability, making it highly suitable for large-scale HMRS. Comprehensive numerical experiments validating these theoretical complexity bounds, particularly the scalability regarding the robot fleet size and types, are presented in Section VI-A.

\section{EXPERIMENTS}

In the experiments section, numerical experiments, simulations, and physical experiments were conducted to demonstrate the performance of the proposed algorithm. All algorithms were implemented using Python 3.10 and ROS Noetic.

\subsection{Numerical Experiments}

In this section, we conduct numerical experiments to validate the proposed methods. Specifically, the experiments are divided into two parts. The first evaluates the effectiveness of the proposed maximum transition edge strategy, and the second assesses the overall performance of the proposed E-PDT framework against an MILP baseline. All experiments were executed within a VMware virtual machine configured with 16GB of RAM and 16 virtual CPU cores, hosted on a workstation equipped with an Intel Core i7-12700K CPU and 32GB of RAM.

\begin{table}[tb]
    \begin{center}
        \caption{Performance comparison of selecting transition edges.}
        \label{tab_ltl_comparison}
        \setlength{\extrarowheight}{2pt}
        \begin{tabular}{c|c|c|c|c}
        \hline
            scLTL & Metric & Method 1 & Method 2 & Method 3 \\ \hline
            \multirow{2}{*}{$\phi_1$} & Cost & \textbf{70} & 103 & 103 \\ 
             & Time(/s) & \textbf{0.1039} & 0.4639 & 0.5278 \\ \hline
            \multirow{2}{*}{$\phi_2$} & Cost & \textbf{69.75} & 83.8 & 79.4 \\ 
             & Time(/s) & \textbf{1.1479} & 3.5324 & 3.7962 \\ \hline
            \multirow{2}{*}{$\phi_3$} & Cost & \textbf{69} & 96.45 & 94.15 \\ 
             & Time(/s) & \textbf{0.2636} & 1.8734 & 2.149 \\ \hline
            \multirow{2}{*}{$\phi_4$} & Cost & \textbf{68.5} & 82.4 & 78.8 \\ 
             & Time(/s) & 1.8326 & \textbf{1.5954} & 1.646 \\ \hline
            \multirow{2}{*}{$\phi_5$} & Cost & \textbf{64.85} & 68.65 & 68.6 \\ 
             & Time(/s) & 0.9385 & \textbf{0.2912} & 0.3167 \\ \hline
            \multirow{2}{*}{$\phi_6$} & Cost & \textbf{68.3} & 68.85 & \textbf{68.3} \\ 
             & Time(/s) & \textbf{0.0564} & 0.1211 & 0.1394 \\ \hline
             \multirow{2}{*}{$\phi_7$} & Cost & \textbf{93.3} & \textbf{93.3} & \textbf{93.3} \\ 
             & Time(/s) & \textbf{0.0304} & 0.0332 & 0.0322 \\ \hline
        \end{tabular}
    \end{center}
\end{table}

We first conduct an experiment to verify the contribution of the maximum transition edge strategy. In this evaluation, we set up three comparative groups, which are selecting only the maximum transition edge (Method 1) and traversing all transition edges in ascending order (Method 2) as well as descending order (Method 3). The tests were conducted across 20 randomly generated environments. The corresponding results are summarized in Table \ref{tab_ltl_comparison}, where $\phi_1 = \lozenge ap_1 \wedge \lozenge ap_2 \wedge \lozenge ap_3$, $\phi_2 = \lozenge (ap_1 \wedge \lozenge (ap_2 \wedge \lozenge ap_3))$, $\phi_3 = \lozenge (ap_1 \wedge \lozenge ap_2) \wedge \lozenge ap_3$, $\phi_4 = \lozenge (ap_1 \wedge \lozenge (ap_2 \vee ap_3))$, $\phi_5 = \lozenge (ap_1 \wedge \lozenge ap_2) \vee \lozenge ap_3$, $\phi_6 = (\neg ap_2 \ \mathcal{U} \ ap_1) \wedge \lozenge ap_2$, and $\phi_7 = \lozenge (ap_1 \wedge \bigcirc (\neg ap_3 \ \mathcal{U} \ ap_2))$.

The results in Table \ref{tab_ltl_comparison} explicitly validate the rationale behind our transition edge selection strategy discussed in Section IV-B, where Method 1 demonstrates fast computation speeds under most scLTL constraints and, crucially, achieves the lowest (or one of the lowest) execution cost across all tested scLTL constraints. In contrast, traversing all transition edges (Method 2 and Method 3) occasionally identifies alternative policies with lower theoretical regret along branches with fewer atomic propositions. However, this lower regret does not necessarily translate into a lower actual execution cost. The empirical results confirm that by prioritizing the maximum transition edge, our framework successfully leverages this information gain to aggressively prune the uncertainty space, thereby reducing the actual execution cost and significantly improving planning efficiency.

\begin{table}[tb]
    \begin{center}
        \caption{Computation time against number of robots.}
        \label{tab1}
        \setlength{\extrarowheight}{2pt}
        \begin{tabular}{c|c|c|c}
        \hline
            $n_a$ & E-PDT(/s) & E-PDT without BnB(/s) & MILP(/s) \\ \hline
            18 & 0.0385 & 0.2745 & 3.4245 \\ 
            30 & 0.0656 & 0.3304 & 5.6970 \\ 
            150 & 0.0915 & 0.5699 & 28.3087 \\ 
            600 & 0.2340 & 1.4592 & 118.4246 \\ 
            1500 & 0.5477 & 3.1825 & 382.2129 \\ 
            3000 & 1.0299 & 6.1049 & 1408.1137 \\ 
            6000 & 1.9625 & 12.0806 & Failed \\ 
            9000 & 3.0045 & 17.5065 & Failed \\ \hline
        \end{tabular}
    \end{center}
\end{table}

\begin{table}[tb]
    \begin{center}
        \caption{Computation time against number of robot types.}
        \label{tab2}
        \setlength{\extrarowheight}{2pt}
        \begin{tabular}{c|c|c|c}
        \hline
            $n_t$ & E-PDT(/s) & E-PDT without BnB(/s) & MILP(/s) \\ \hline
            3 & 0.0506 & 0.3534 & 18.9809 \\ 
            10 & 0.0669 & 0.4472 & 63.6334 \\ 
            30 & 0.1437 & 0.7990 & 126.7416 \\ 
            50 & 0.2323 & 1.2010 & 423.1917 \\ 
            60 & 0.2869 & 1.5743 & 7665.2873 \\ 
            80 & 0.3510 & 1.9824 & Filed \\ 
            100 & 0.4286 & 2.3987 & Failed \\ \hline
        \end{tabular}
    \end{center}
\end{table}

Subsequently, we compare the performance of the E-PDT framework against the MILP baseline \cite{luo2022temporal}. For this specific comparison, to represent decision-making under the worst-case scenario, the MILP baseline is configured to plan exclusively toward guaranteed regions. Furthermore, to ensure fairness, we employ only the high-level planning component of the baseline, excluding fine-grained trajectory planning for collision avoidance.
 
Regarding the experimental setup, we employ a fixed scLTL formula $\phi = \lozenge ap_1 \wedge \lozenge ap_2 \wedge \lozenge ap_3$ and a constant environmental scale. First, we fix the number of robot types at 3 and vary the fleet size of each type from 6 to 3,000, while keeping other parameters constant; the resulting computation time is illustrated in Table \ref{tab1}. Subsequently, we fix the fleet size at 10 robots per type and vary the number of robot types from 3 to 100, with the types evenly distributed among the three RbAPs. The corresponding planning time is presented in Table \ref{tab2}.

The experimental results demonstrate that the computation time of E-PDT scales linearly or near-linearly with respect to the number of robots and types, which aligns with our theoretical analysis. In contrast, the complexity of the MILP approach increases drastically with the problem size, eventually failing to generate a solution. It is worth noting that E-PDT solves the Min-Max Regret problem, which is theoretically more demanding than the problem solved by the MILP baseline, as it requires computing the hindsight-optimal cost for potential outcomes. Despite this higher complexity class, E-PDT still achieves orders-of-magnitude faster solution times, highlighting the efficiency of the regret-based pruning over traditional integer programming.

\subsection{Simulation Experiments}

\begin{figure*}[htbp]
	\centering
    \includegraphics[width=1.0\linewidth]{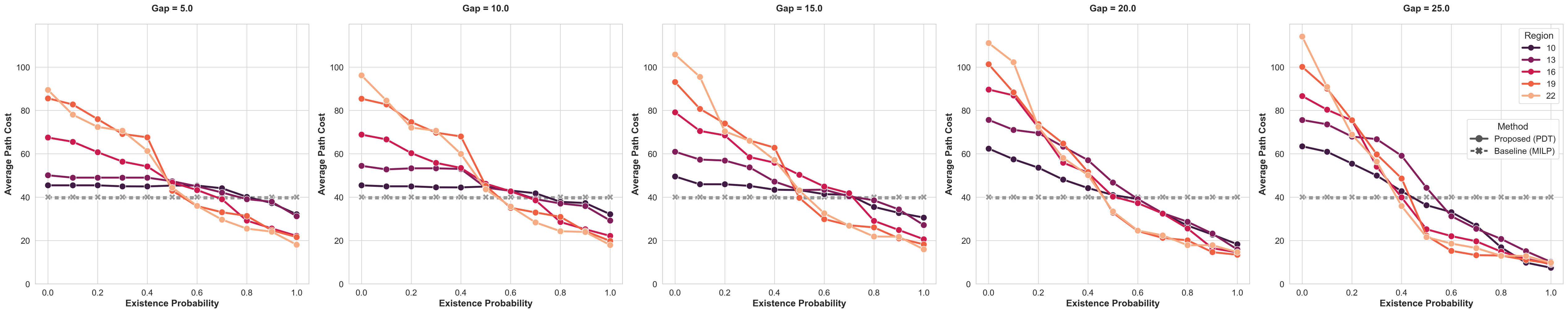}
    \caption{Cost against potential regions probability with fixed gap.}
    \label{vary_gap}
\end{figure*}

\begin{figure*}[htbp]
	\centering
    \includegraphics[width=1.0\linewidth]{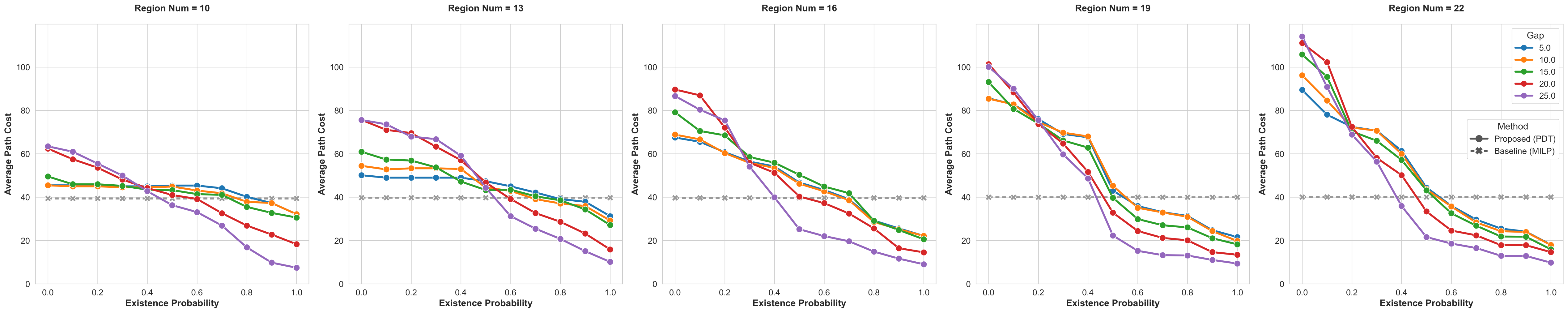}
    \caption{Cost against potential regions probability with fixed number of regions.}
    \label{vary_region}
\end{figure*}

In this section, we conducted extensive simulation experiments to evaluate the task completion cost of the proposed E-PDT framework compared to the MILP baseline under varying conditions of environmental complexity and uncertainty. The MILP settings and the experimental hardware configuration remain consistent with the numerical experiments section.

\begin{figure}[tb]
      \centering
      \includegraphics[width=0.49\textwidth]{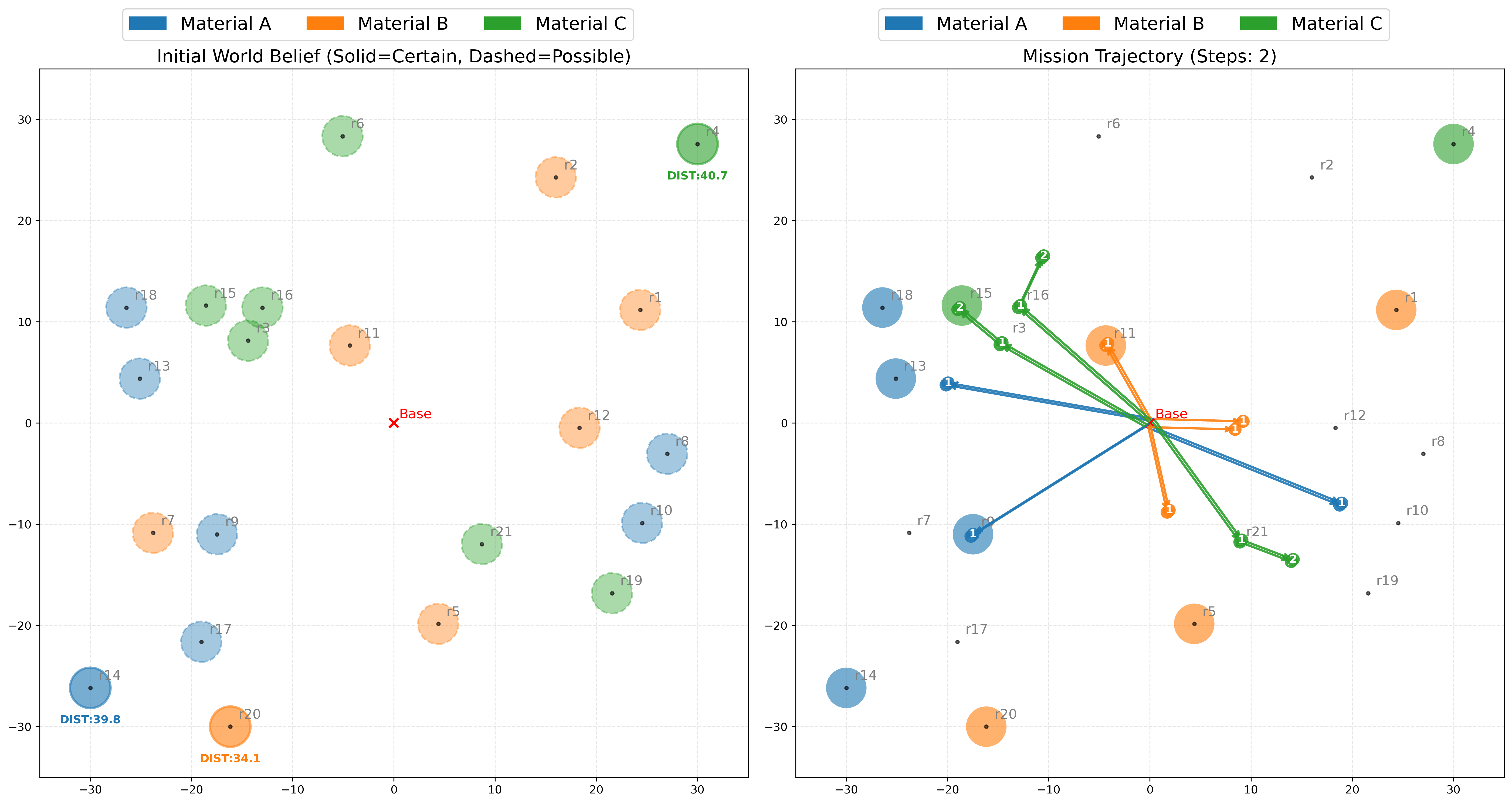}
      \caption{An example of a random world (left) and corresponding result trajectory (right). In the left figure, different color represents different material in initial world belief. The certain region is solid, while the potential region is dashed. In the right figure, the pointed number represents the mission step.}
      \label{env+traj}
\end{figure}

The simulation environment is established within a $60 \times 60$ continuous workspace, as shown in Fig. \ref{env+traj} left. The mission is executed by a heterogeneous fleet consisting of 3 distinct robot categories, with 2 robots per category (resulting in a total system size of 6). The task specification scLTL formula is $\phi = \lozenge ap_1 \wedge \lozenge ap_2 \wedge \lozenge ap_3$. Crucially, the satisfaction of each RbAP requires the collaborative effort of 2 robots of the corresponding type, which imposes a concurrency constraint allowing only one group of robots to execute a specific task type at any given time. We simulated diverse environmental conditions by systematically varying three parameters: the minimum distance between deterministic and potential regions (Gap $\in [5, 25]$), the total number of regions ($|R| \in [10, 22]$), and the existence probability of resources in potential regions ($p \in [0, 1]$). For each parameter combination, 20 randomized seeds were conducted to eliminate stochastic variance, and the average path cost was recorded. Fig. \ref{vary_gap} illustrates the cost evolution with respect to probability across different region counts under fixed minimum distances, while Fig. \ref{vary_region} depicts the cost trends across different minimum distances for fixed region counts. 

Experimental results indicate that in scenarios with low resource probability, the E-PDT incurs a Robustness Premium. As environmental uncertainty decreases (i.e., the probability of existence increases), the benefit of exploration outweighs the risk, and the overall cost of E-PDT exhibits a downward trend, eventually outperforming the baseline. This performance gap fundamentally reflects the structural limitation of the worst-case metric used in the MILP baseline. By optimizing strictly for the worst-case scenario, the baseline adopts an overly conservative strategy that effectively prohibits the exploration of potential regions. In contrast, the min-max regret metric effectively captures and exploits the increasing richness of environmental resources.

We define the critical threshold at which the E-PDT cost becomes superior to the MILP cost as the Crossover Probability. As observed in the Fig. \ref{vary_gap} and Fig. \ref{vary_region}, this crossover probability exhibits a left-shift trend as environmental complexity increases. Specifically, when the minimum distance is set to 20, the crossover probability approaches 0.5 across almost all region counts, corresponding to the state of maximum entropy. Furthermore, in more complex environments, this threshold drops to 0.4 or even lower. These results demonstrate that while worst-case planning is valid in extreme scarcity, regret-based planning is structurally superior in PKE where exploration yields significant efficiency gains.

\begin{figure}[tb]
      \centering
      \includegraphics[width=0.49\textwidth]{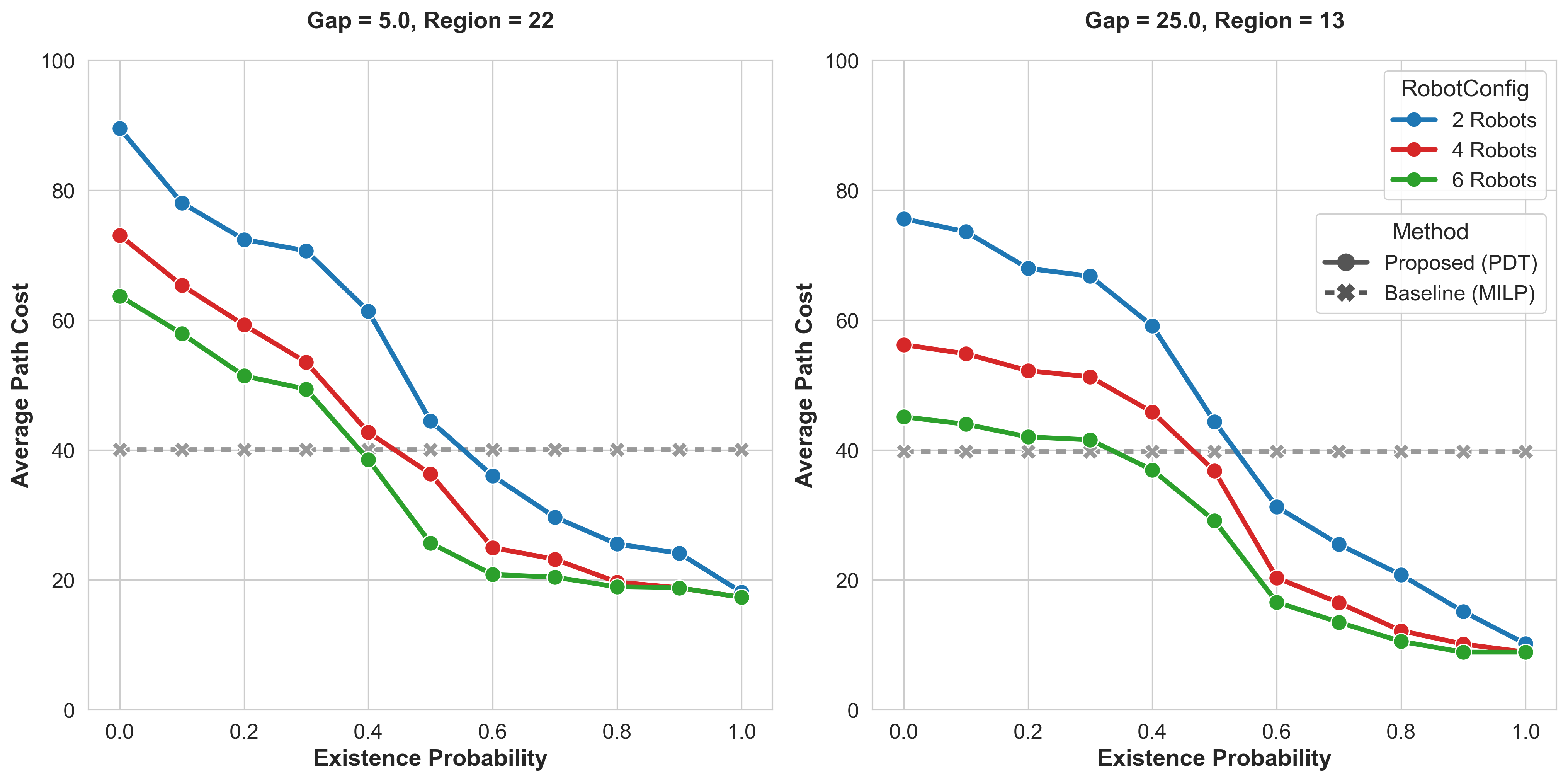}
      \caption{Cost against number of robot types.}
      \label{vary_robot}
\end{figure}

To further assess scalability, we extended the experiments by increasing the number of robots per category to 4 and 6, the result is shown in Fig. \ref{vary_robot}. Fig. \ref{env+traj} (right) illustrates the robot trajectories for an experiment involving six robots of each type. It can be observed that the generated plan unfolds in two distinct phases. In the first phase, type I robots are dispatched to regions $r_9$, $r_{10}$, and $r_{13}$, respectively. The sub-group assigned to $r_9$ arrives first and successfully discovers resource $A$, which triggers the immediate termination of tasks for the remaining sub-groups. A similar scenario occurs with the type II robot fleet. In contrast, the type III robots fail to discover resource $C$ in the three regions visited during the first phase and thus continue the plan. The entire mission concludes once a sub-group discovers resource $C$ in region $r_{15}$, prompting the termination of the entire mission. The results demonstrate that our algorithm can effectively leverage the increased scale of the robot fleet. By optimizing resource allocation, E-PDT utilizes the additional robots to conduct concurrent exploration and exploitation, thereby enhancing mission efficiency even in highly uncertain environments.

\subsection{Physical Experiments}

To evaluate the proposed framework, physical experiments are conducted using a multi-robot platform of Wheeltec R550 mobile robots. Each robot utilizes an Intel NUC 12 for real-time onboard computation and an OptiTrack system for global localization. The software architecture is implemented in ROS, with real-time inter-robot communication facilitated via a WiFi network.

\begin{figure}[tb]
      \centering
      \includegraphics[width=0.49\textwidth]{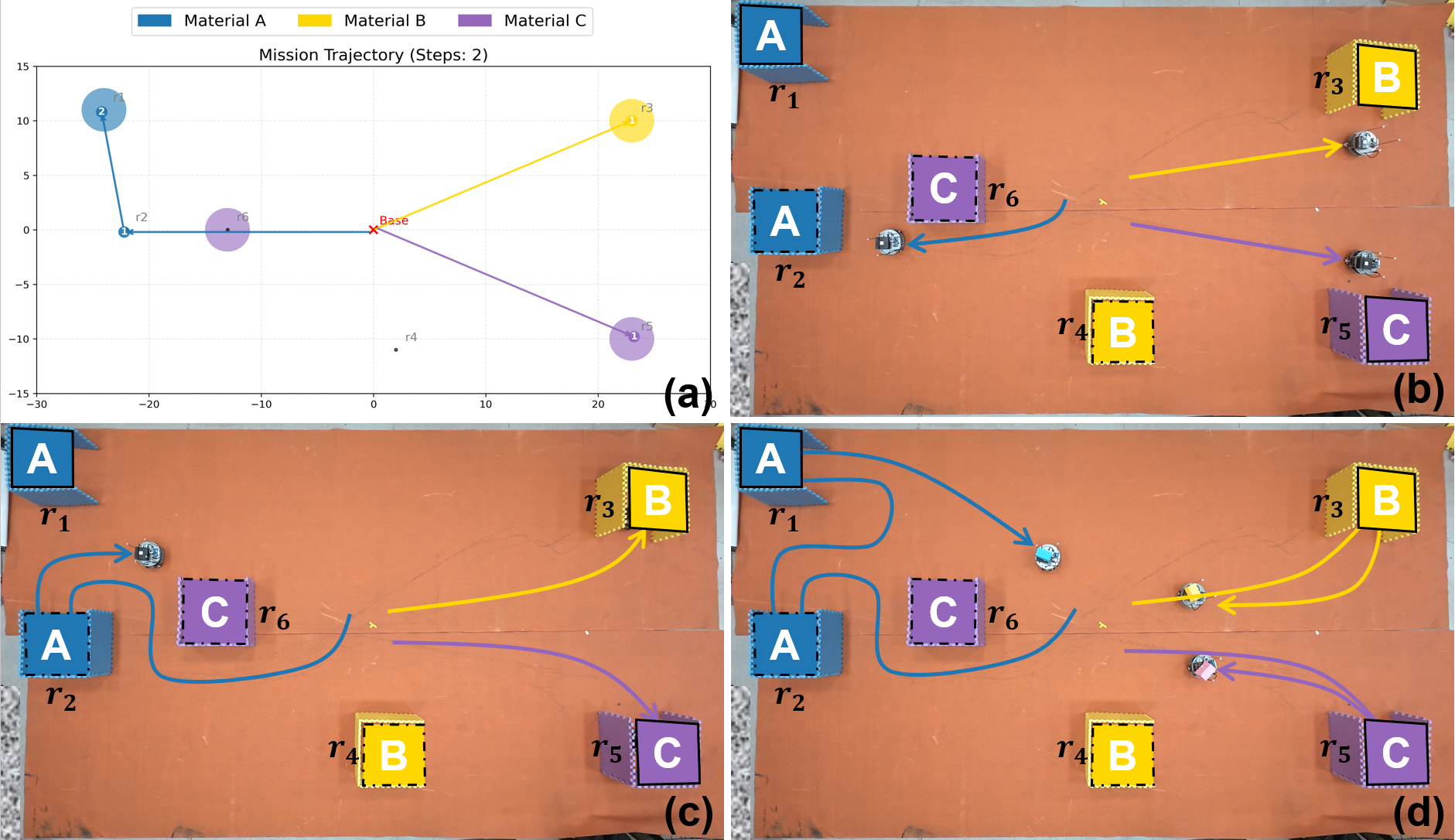}
      \caption{The result of physical experiment 1. Solid boundaries denote guaranteed regions, while dashed boundaries indicate potential regions. In (a), the colored circular markers represent the ground truth, indicating the regions where the resources actually exist.}
      \label{phy_exp1}
\end{figure}

In the first experiment, as shown in Fig. \ref{phy_exp1}, we deployed a heterogeneous fleet of three robots ($n_t=3$) across 6 regions to collect 3 distinct resource types and return to the base. Each resource type was associated with one guaranteed region and one potential region. The task is formally expressed as $\phi = (\lozenge ap_1 \wedge \lozenge ap_2 \wedge \lozenge ap_3) \wedge \lozenge ap_{base}$. Our algorithm generated the policy tree in 0.0065 seconds. The initial policy dispatched robot $a_1$ to explore a potential region, while $a_2$ and $a_3$ were sent directly to the guaranteed regions. During execution, $a_1$ failed to discover resource $A$ in the potential region $r_2$; consequently, the contingent policy redirected it to the guaranteed region $r_1$. Finally, all three robots returned to the base upon task completion.

\begin{figure}[tb]
      \centering
      \includegraphics[width=0.49\textwidth]{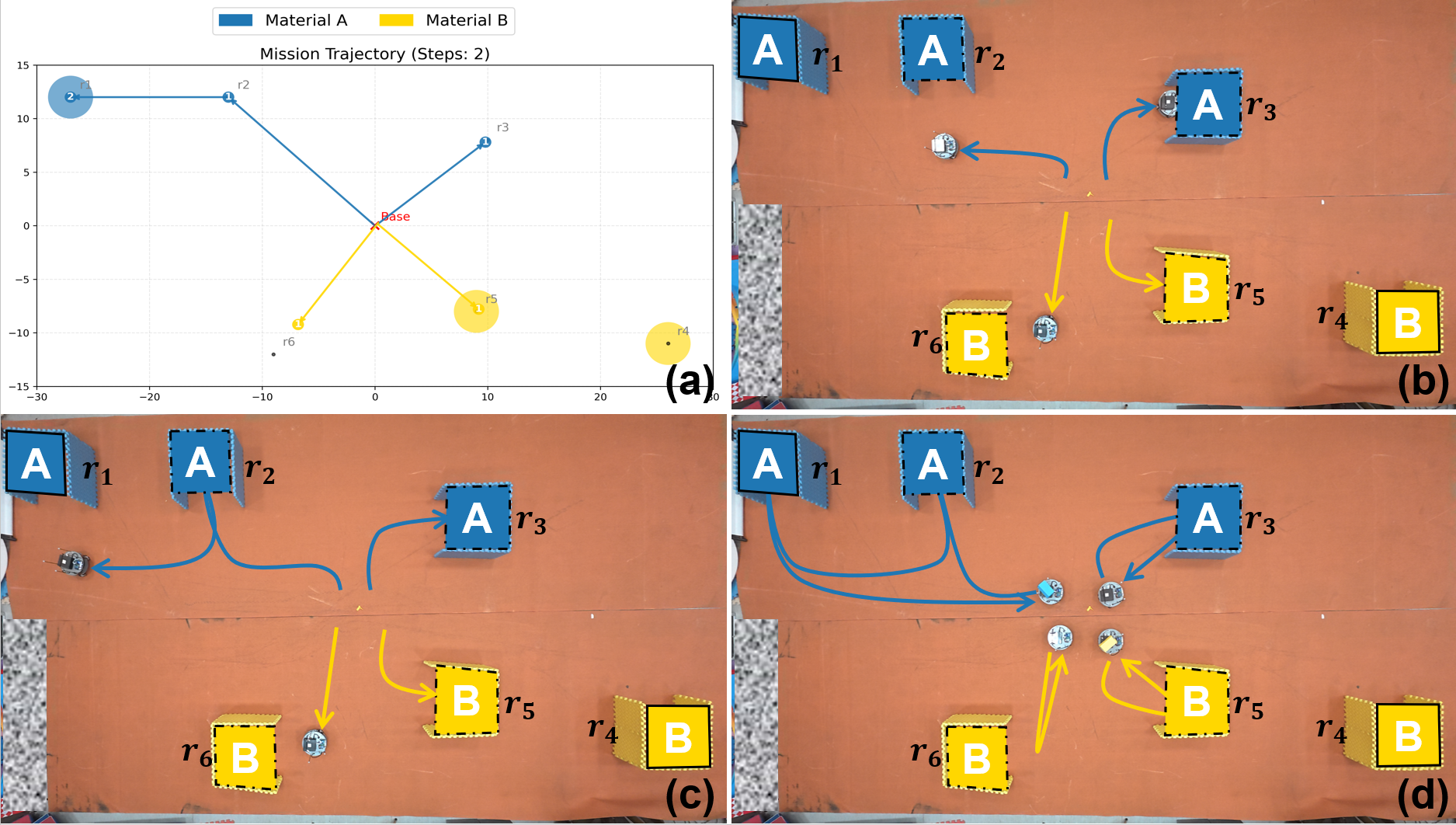}
      \caption{The result of physical experiment 2. The notation used is the same as in Fig. \ref{phy_exp1}.}
      \label{phy_exp2}
\end{figure}

In the second experiment, we utilized four robots belonging to two distinct types to collect 2 resource types across 6 regions and return to the base. The environment was configured such that each resource had one guaranteed region and two potential regions. The proposed algorithm generated the policy tree in 0.0078 seconds, adopting a strategy that dispatched all four robots to prioritize exploration. During the operation, both $a_1$ and $a_2$ failed to detect resource $A$ in their respective potential regions; therefore, the policy redirected $a_1$ to the guaranteed region $r_1$. Concurrently, robot $a_4$ successfully discovered resource $B$ in region $r_5$ and communicated this information to the team. Upon receiving the update, robot $a_3$ immediately halted its search. The mission concluded after $a_1$ secured resource A in $r_1$, and all robots returned to the base.

\section{CONCLUSIONS}

This paper proposes the E-PDT framework, combining RbAP and a min-max regret pruning strategy, to efficiently solve temporal logic task planning problems for large-scale HMRS in PKE. Both theoretical and experimental results indicate that our approach effectively balances exploration and exploitation. Compared to MILP baselines, it achieves near-linear scalability and superior solution quality. 

It is worth noting that the current framework focuses on formal modeling and planning for predefined environmental uncertainties. It is not currently designed to handle model-external disturbances or dynamic changes that are not explicitly enumerated in advance. Furthermore, the current formal abstraction relies on relatively strong structural assumptions that may be challenging to satisfy in highly unstructured scenarios. Future work will focus on extending the framework to fully dynamic and unknown environments by integrating reactive synthesis mechanisms, as well as exploring learning methods for the modeling of complex, unstructured uncertainties, to facilitate robust decision-making in more complex tasks.

\bibliographystyle{IEEEtran}
\bibliography{Citation}

 
\vspace{11pt}

\begin{IEEEbiography}[{\includegraphics[width=1in,height=1.25in,clip,keepaspectratio]{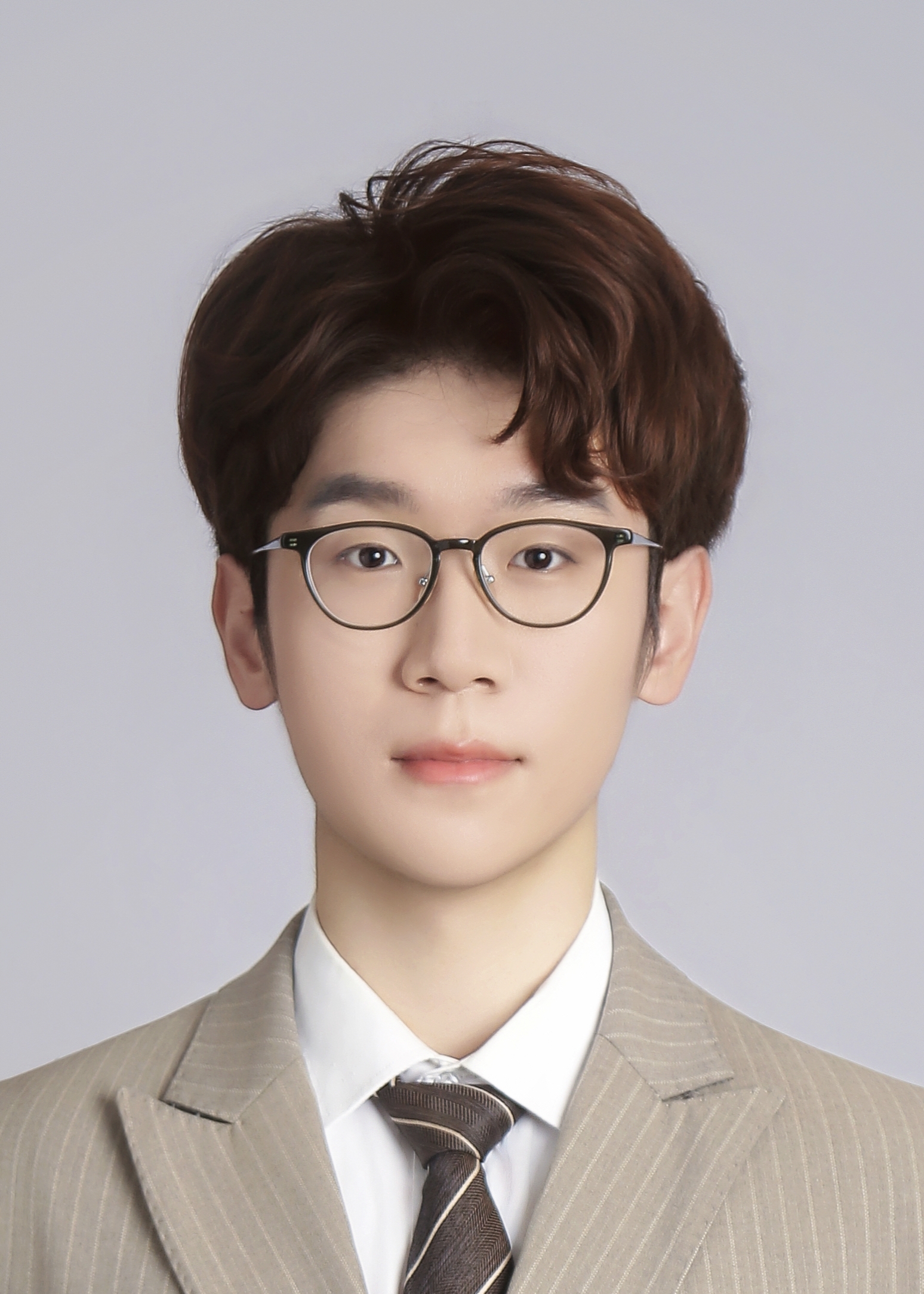}}]{Xinkai Liang} received the B.S. degree in automation in 2023 from Beijing Institute of Technology, Beijing, China, and is currently pursuing the Ph.D. degree with the School of Automation, Beijing Institute of Technology, Beijing, China. His research interests include formal methods, temporal logic, multi-agent systems, task and motion planning.
\end{IEEEbiography}

\begin{IEEEbiography}[{\includegraphics[width=1in,height=1.25in,clip,keepaspectratio]{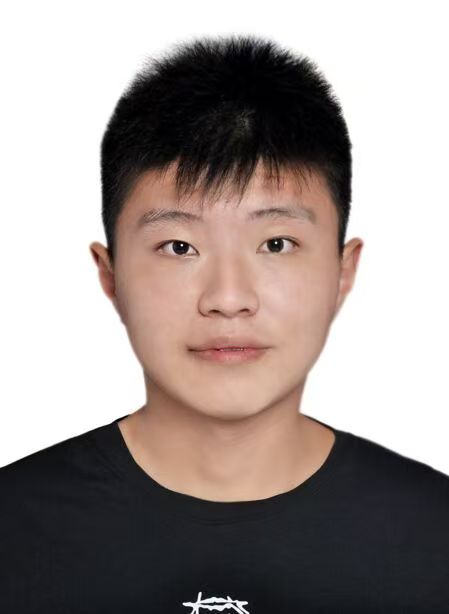}}]{Huixuan Chan} is currently pursuing the Ph.D. degree at Beijing Institute of Technology, Beijing, China. His research interests focus on linear temporal logic–based multi-robot task allocation and opponent intention prediction using contingency games.
\end{IEEEbiography}

\begin{IEEEbiography}[{\includegraphics[width=1in,height=1.25in,clip,keepaspectratio]{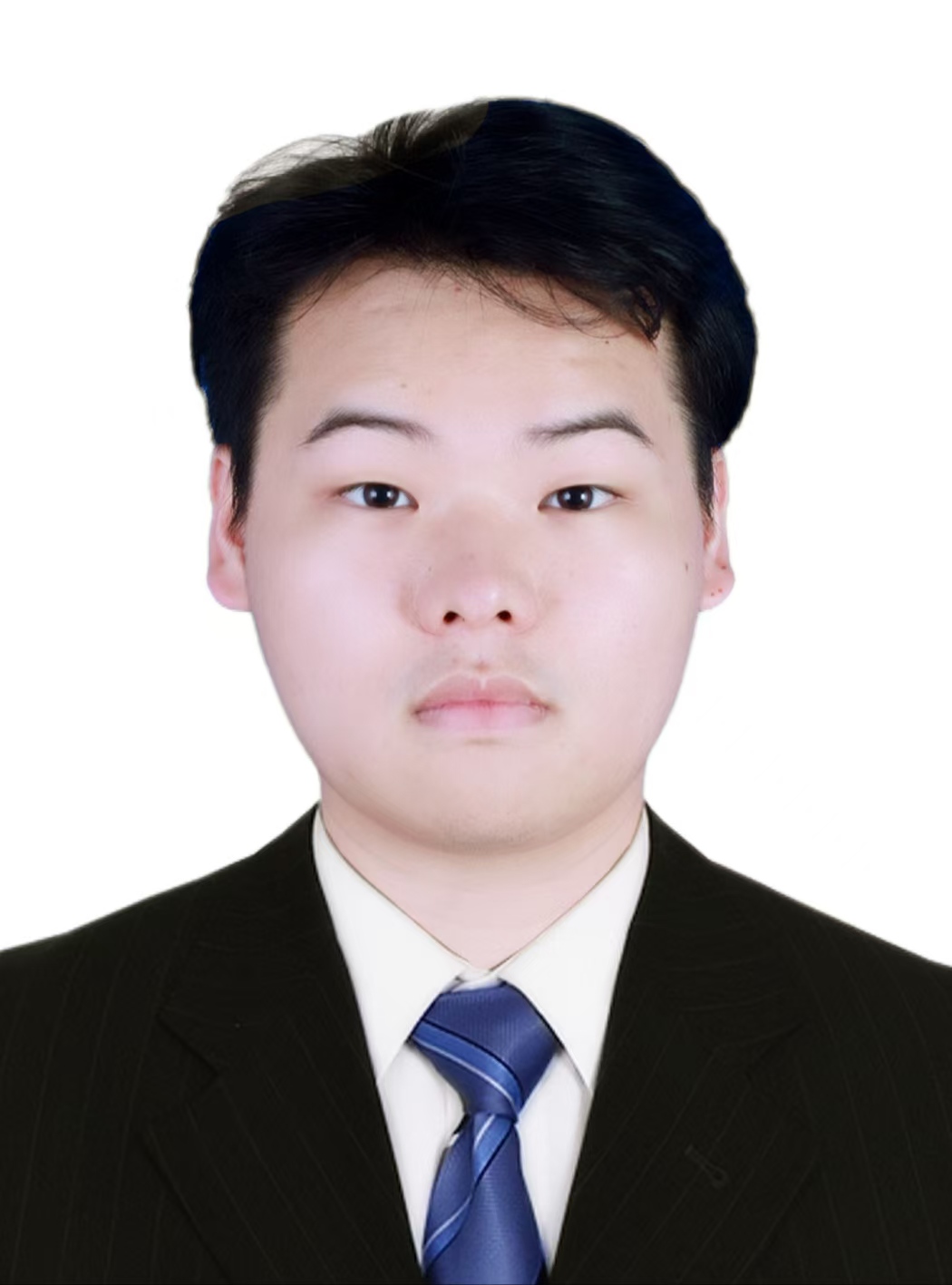}}]{Ying Liu} received the B.S. degree in robotics engineering, in 2024, from the University of Science and Technology Beijing, Beijing, China. He is currently working toward the M.S. degree in control science and engineering in Beijing Institute of Technology. His research interests include control of multi-agent systems and mobile robots.
\end{IEEEbiography}

\begin{IEEEbiography}[{\includegraphics[width=1in,height=1.25in,clip,keepaspectratio]{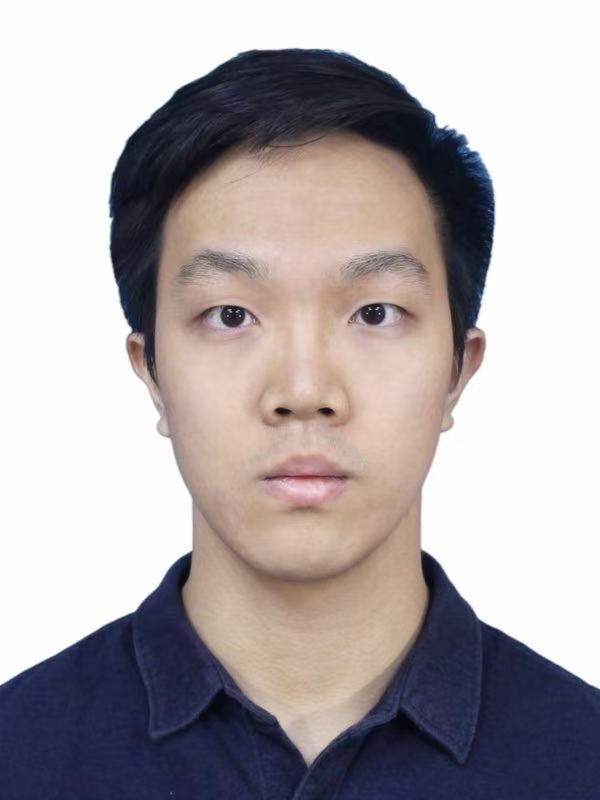}}]{Yangxi Shi} is currently pursuing the Ph. D. degree with the School of Automation, Beijing Institute of Technology, Beijing, China. His research interests include UAV-Swarm perception, cooperative planning and control.
\end{IEEEbiography}

\begin{IEEEbiography}[{\includegraphics[width=1in,height=1.25in,clip,keepaspectratio]{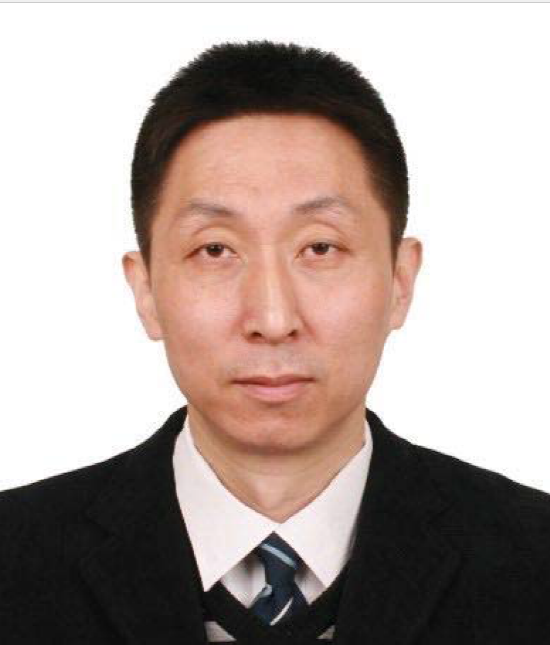}}]{Hao Fang} (Member, IEEE) received the B.S. degree from the Xi’an University of Technology, Shaanxi,
China, in 1995, and the M.S. and Ph.D. degrees from the Xi’an Jiaotong University, Shaanxi, in 1998 and
2002, respectively. Since 2011, he has been a Professor with the Beijing Institute of Technology, Beijing, China. He held two postdoctoral appointments with the INRIA/France Research Group of COPRIN
and the LASMEA (UNR6602 CNRS/Blaise Pascal University, Clermont-Ferrand, France). His research interests include all-terrain mobile robots, robotic control, and multi-agent systems.
\end{IEEEbiography}

\vspace{11pt}

\vfill

\end{document}